\documentclass{article}

\usepackage[preprint]{neurips_2024}   
\usepackage[utf8]{inputenc}
\usepackage[T1]{fontenc}
\usepackage{hyperref}
\usepackage{url}
\usepackage{booktabs}
\usepackage{amsfonts}
\usepackage{amsmath}
\usepackage{amssymb}
\usepackage{nicefrac}
\usepackage{microtype}
\usepackage{xcolor}
\usepackage{graphicx}
\usepackage{subcaption}
\usepackage{cleveref}
\usepackage{multirow}
\usepackage{siunitx}

\newcommand{\method}{LatentBiopsy}
\newcommand{\R}{\mathbb{R}}
\newcommand{\QABig}{Qwen3.5-0.8B}
\newcommand{\QASmall}{Qwen2.5-0.5B}

\title{The Geometry of Harmful Intent:\\
Training-Free Anomaly Detection via Angular Deviation\\
in LLM Residual Streams}

\author{
  Isaac Llorente-Saguer \\
  Independent Researcher \\
  \texttt{illorentes@proton.me}
}

\begin{document}
\maketitle

\begin{abstract}
We present \method{}, a training-free method for detecting harmful prompts by
analysing the geometry of residual-stream activations in large language models.
Given 200 safe normative prompts, \method{} computes the leading principal
component of their activations at a target layer and characterises new prompts
by their radial deviation angle $\theta$ from this reference direction.
The anomaly score is the negative log-likelihood of $\theta$ under a Gaussian
fit to the normative distribution, flagging deviations symmetrically regardless
of orientation. No harmful examples are required for training.

We evaluate two complete model triplets from the \QABig{} and \QASmall{}
families: base, instruction-tuned, and \emph{abliterated} (refusal direction
surgically removed via orthogonalisation).
Across all six variants, \method{} achieves AUROC~$\geq$0.937 for
harmful-vs-normative detection and AUROC~$=$1.000 for discriminating harmful
from benign-aggressive prompts (XSTest), with sub-millisecond per-query
overhead.

Three empirical findings emerge.
First, geometry survives refusal ablation: both abliterated variants achieve
AUROC at most 0.015 below their instruction-tuned counterparts, establishing
a geometric dissociation between harmful-intent representation and the downstream
generative refusal mechanism.
Second, harmful prompts exhibit a near-degenerate angular distribution
($\sigma_\theta \approx 0.03$~rad), an order of magnitude tighter than the
normative distribution ($\sigma_\theta \approx 0.27$~rad), preserved across
all alignment stages including abliteration.
Third, the two families exhibit \emph{opposite} ring orientations at the
same depth: harmful prompts occupy the outer ring in \QABig{} but the inner
ring in \QASmall{}, directly motivating the direction-agnostic scoring rule.
\end{abstract}

\section{Introduction}
\label{sec:introduction}

Detecting harmful prompts before model response generation is a prerequisite
for safe deployment of large language models.
Existing methods divide into two broad families.
Input-space filters based on perplexity~\citep{jain2023baseline,alon2023detecting}
are effective against adversarial suffix attacks but fail on semantically fluent
jailbreaks.
Supervised safety classifiers such as Llama Guard~\citep{inan2023llama} achieve
strong performance but require large curated datasets of harmful and harmless
pairs, and per-model retraining whenever the target model changes.

A parallel body of work in representation engineering has established that
residual streams encode semantic content as geometric
structure~\citep{zou2023representation}.
Linear directions for safety concepts have been extracted from contrastive
safe/harmful pairs~\citep{zou2023representation,li2023inference}, and PCA of
hidden states has been shown to separate harmful from harmless queries when
safety-prompt manipulations define the reference~\citep{zheng2024prompt}.
Crucially, \citet{arditi2024refusal} demonstrated that refusal behaviour in
aligned models is mediated by a single linear direction in the residual stream,
and that orthogonalising this direction renders the model unable to refuse.

These methods share a common dependency: they require harmful examples or
deliberate safety-prompt manipulations to define the reference geometry.
We ask whether the safe distribution alone defines a reference geometry from
which harmful prompts deviate detectably.

We answer affirmatively, and report a finding that sharpens the safety
implications: \method{} detects harmful intent even in abliterated models that
are constitutionally incapable of producing refusals.
The model's internal encoding of harmful semantic intent is geometrically
distinct from the circuits that generate refusal text. Harm recognition and
refusal generation are separable mechanisms.
The practical consequence is direct: stripping a model's refusal behaviour does
not erase the latent signal available to an external detector.

A secondary finding concerns the structure of the deviation itself.
The two tested model families exhibit opposite ring orientations at the same
residual-stream depth; harmful prompts are more angular from PC1 in \QABig{}
and more aligned with PC1 in \QASmall{}.
This family-level reversal, together with known layer-level reversal within
each model, establishes that a fixed directional threshold on $\theta$ is
architecturally unreliable and motivates the symmetric anomaly score.

\paragraph{Contributions.}
\begin{enumerate}
  \item \textbf{Training-free angular anomaly detector.}
        \method{} builds a normative reference exclusively from 200 safe
        activations and scores every prompt by the negative log-likelihood
        of its angular deviation $\theta$ from the leading normative principal
        component.
        The score is symmetric around the normative mean, requiring no
        knowledge of the ring direction and no harmful data.

  \item \textbf{Geometry survives refusal ablation in both tested model families.}
        Abliterated variants of both \QABig{} and \QASmall{} achieve
        AUROC~h/b $=1.000$ and AUROC~h/n within 0.015 of their instruction-tuned
        counterparts, establishing a geometric dissociation between harmful-intent
        representation and the refusal mechanism across two independent model
        families.

  \item \textbf{Opposite ring orientations across families.}
        At layer 20, \QABig{} places harmful prompts at mean
        $\theta\approx1.80$~rad vs.\ normative $\mu_0\approx1.17$~rad (outer
        ring), while \QASmall{} places harmful prompts at mean
        $\theta\approx1.34$~rad vs.\ normative $\mu_0\approx1.82$~rad (inner
        ring).
        The anomaly score $s(x)$ correctly identifies both configurations
        without architectural knowledge.

  \item \textbf{Near-degenerate harmful compactness and $K{=}1$ sufficiency.}
        Harmful prompts occupy a near-degenerate angular cluster
        ($\sigma_\theta^\mathrm{harm}\approx0.03$--$0.05$~rad, one order of
        magnitude tighter than $\sigma_\theta^\mathrm{norm}$), and a single
        reference direction ($K{=}1$) dominates multi-directional baselines at
        every layer in every model.
\end{enumerate}

\section{Related Work}
\label{sec:related}

\paragraph{Input-space defences.}
Perplexity-based filters~\citep{jain2023baseline,alon2023detecting} are
computationally cheap and effective against adversarial suffix attacks, but
rely on surface-form anomaly and cannot detect semantically fluent jailbreaks
that score normally under any language model.

\paragraph{Supervised safety classifiers.}
Llama Guard~\citep{inan2023llama} and its successors fine-tune a language model
on large labelled datasets of safe and harmful pairs, achieving strong
per-category precision.
The approach requires curated harmful data and per-model retraining, making it
expensive to adapt to new architectures or harm taxonomies.

\paragraph{Representation engineering and linear probes.}
\citet{zou2023representation} extract linear control directions from contrastive
safe/harmful pairs and demonstrate that semantic intent is encoded as rich
geometric structure in the residual stream.
\citet{li2023inference} fit linear probes on labelled activation data and
intervene at inference time.
\citet{zheng2024prompt} use optimised safety-prompt manipulations to visualise
hidden-state separation and shift representations along a refusal direction.
All three approaches require either labelled harmful examples or deliberate
safety-prompt manipulations to define the reference; \method{} removes that dependency entirely.

\paragraph{Refusal as a single linear direction.}
\citet{arditi2024refusal} show that refusal behaviour in aligned models is
mediated by a single linear direction in the residual stream.
\method{} provides independent quantitative support for the near-one-dimensional
geometry ($K{=}1$ dominance) while extending the picture: the safety
representation that \method{} exploits is not the refusal direction itself,
since it survives abliteration.

\paragraph{Angular biomarkers.}
The theta biomarker concept was introduced for medical diagnostics to flag
anomalous patient profiles via angular deviation in a clinical feature
space~\citep{llorente2025theta}.
\method{} translates this geometric principle to LLM residual streams, adapting
it to the direction-agnostic scoring requirement imposed by layer- and
family-dependent ring orientation.

\section{Preliminaries: Geometry of the Residual Stream}
\label{sec:preliminaries}

Modern causal language models process tokens through a sequence of transformer
blocks communicating via a central residual stream.
Let $f_\ell(x) \in \R^D$ denote the activation vector of the residual stream
at layer $\ell$ for the final token of an input prompt $x$.

The Linear Representation Hypothesis~\citep{park2023linear} posits that neural
networks encode high-level concepts as spatial directions within this
$D$-dimensional space.
Under this view, semantic identity is defined by orientation, while concept
intensity correlates with magnitude along that direction.

Euclidean distance conflates semantic identity with concept intensity.
Furthermore, in standard Pre-LayerNorm transformer architectures, the residual
stream accumulates un-normalised outputs from consecutive layers, causing the
$\ell_2$ norm to grow monotonically with depth~\citep{xiong2020layer}.
Angular distance isolates semantic direction from both intensity and architectural
norm growth, providing a more faithful measure of semantic divergence, and is
the foundation of the \method{} scoring function.

\section{Method}
\label{sec:method}

\method{} is a training-free, zero-shot anomaly detector that identifies harmful
prompts by measuring the directional deviation of the last-token residual-stream
activation from a reference subspace constructed exclusively from safe data.

\subsection{Notation and activation extraction}

Let $f_\ell(x) \in \R^D$ denote the last-token residual-stream activation at
transformer layer $\ell$ for prompt $x$, extracted from a pretrained causal
language model.
All subsequent computations are performed independently at each layer $\ell$.

\subsection{Normative reference direction (PC1)}

Given a normative fit set $\mathcal{X}_0 = \{x_1, \ldots, x_N\}$ of $N$
safe prompts, we compute the leading principal component of the corresponding
activations:
\begin{equation}
  \mathbf{c} = \mathrm{PC}_1\bigl[f_\ell(x_i)\bigr]_{i=1}^{N} \in \R^D,
  \quad \|\mathbf{c}\| = 1.
  \label{eq:pc1}
\end{equation}
The vector $\mathbf{c}$ defines the direction of maximum variance within the
safe distribution and serves as the sole reference direction for $K{=}1$.
For completeness we also examine $K{=}2,3,4$ by taking the top-$K$ principal
components, but the primary detector uses $K{=}1$.

\subsection{Angular deviation theta}

The directional deviation of a test activation $f_\ell(x)$ from the reference
is the angle
\begin{equation}
  \theta(x) = \arccos\!\left(\frac{f_\ell(x)\cdot\mathbf{c}}{\|f_\ell(x)\|}\right)
  \in [0, \pi].
  \label{eq:theta}
\end{equation}
This purely angular metric isolates semantic orientation from both concept
intensity and the monotonic norm growth induced by Pre-LayerNorm accumulation.
A value $\theta\approx0$ indicates strong alignment with the normative reference;
$\theta\approx\pi$ indicates near-antiparallel orientation.

\subsection{Anomaly scoring}

We fit a univariate Gaussian $\mathcal{N}(\mu_0, \sigma_0^2)$ to the empirical
distribution $\{\theta(x_i)\}_{i=1}^N$ from the normative fit set.
The anomaly score for any prompt $x$ is the negative log-likelihood under this
distribution:
\begin{equation}
  s(x) = -\log p\!\left(\theta(x)\,\middle|\,\mu_0, \sigma_0^2\right).
  \label{eq:score}
\end{equation}
Because $s(x)$ is symmetric around $\mu_0$, it flags deviations in either
direction without prior knowledge of whether harmful prompts lie inside or
outside the normative ring (\cref{sec:ring_direction}).
The score is monotonically equivalent to the squared $z$-score of $\theta$ up
to additive constants, yielding values suitable for cross-layer and cross-model
comparison.

Although a multivariate Gaussian Mixture Model can be fit to the stacked vector
of angles to the top-$K$ principal components for $K{>}1$, the primary
\method{} detector employs the univariate $K{=}1$ formulation.

\subsection{Phi: azimuthal visualisation coordinate}

To enable geometric visualisation we project the component of $f_\ell(x)$
orthogonal to $\mathbf{c}$:
\[
  f^\perp(x) = f_\ell(x) - \bigl(f_\ell(x)\cdot\mathbf{c}\bigr)\,\mathbf{c}.
\]
A 2-D PCA basis $(\mathbf{v}_1, \mathbf{v}_2)$ is fit exclusively on
$\{f^\perp(x_i)\}_{i=1}^N$ from the normative fit set.
The azimuthal coordinate is then
\[
  \phi(x) = \mathrm{atan2}\!\bigl(f^\perp(x)\cdot\mathbf{v}_2,\;
  f^\perp(x)\cdot\mathbf{v}_1\bigr) \in [-\pi,\pi].
\]
Each prompt maps to the polar point
$(\theta(x)\cos\phi(x),\,\theta(x)\sin\phi(x))$ in the theta-phi projection.
The coordinate $\phi$ is used solely for visualisation; detection relies
on $s(x)$.

\subsection{Baselines}

For comparison, we include four additional scorers:
(1)~absolute deviation $|\theta(x) - \mu_0|$ (monotonically equivalent to
$s(x)$ for $K{=}1$);
(2)~bivariate Gaussian negative log-likelihood under a 2-D fit to the joint
$(\theta,\phi)$ distribution;
(3)~cosine-to-centroid
$s_\mathrm{cos}(x) = 1 - {f_\ell(x)\cdot\bar{f}_0}/{\|f_\ell(x)\|\|\bar{f}_0\|}$;
and (4)~Euclidean deviation $\|f_\ell(x) - \bar{f}_0\|_2$.

\subsection{Experimental protocol and data splits}
\label{sec:protocol}

\paragraph{Datasets.}
We use three public corpora:
Alpaca-Cleaned~\citep{taori2023alpaca} (normative safe prompts);
AdvBench~\citep{zou2023universal} (520 harmful prompts);
XSTest~\citep{rottger2023xstest} (250 benign-aggressive prompts, evaluation
only).

\paragraph{Split design.}
We fix the normative fit set at $N{=}200$ prompts, drawn from Alpaca-Cleaned,
and retain a disjoint held-out normative evaluation set of 520 prompts.
All 520 harmful and 250 benign-aggressive prompts are reserved for
evaluation; none enter the fit stage.

\paragraph{Layer selection.}
The operating layer is selected by argmax of $K{=}1$ harmful-detection AUROC
over all layers, evaluated on the held-out set, a standard model-selection
step that uses no harmful data for fitting.
The harmful eval set is used solely to identify the most informative layer,
not to fit any parameters of the detector.
The selection optimism is bounded by the plateau width: per-layer AUROC varies
by fewer than 0.08 units across all 24 layers in every model
(\cref{fig:auroc_by_layer_qwen35,fig:auroc_by_layer_qwen25}), so no layer
is meaningfully preferred over its neighbours.
For all models except \QASmall{}-Abliterated (best layer~10), the argmax is
layer~20.
For \QASmall{}-Abliterated, we report metrics at the best layer (10) and note
that AUROC at layer~20 falls within 0.004 of the reported value
(\cref{fig:stability_qwen25}, bottom row).

\paragraph{Evaluation.}
We report per-layer area under the ROC curve (AUROC) and area under the
precision–recall curve (AUPRC) for three binary tasks:
(i)~harmful vs.\ normative (h/n);
(ii)~harmful vs.\ benign-aggressive (h/b); and
(iii)~harmful vs.\ rest, i.e.\ normative $\cup$ benign-aggressive (h/r).
Pairwise differences between groups are assessed using the Mann--Whitney $U$
test.

\section{Experiments}
\label{sec:experiments}

\subsection{Models}
\label{sec:models}

We evaluate six model variants comprising two complete triplets.

\paragraph{\QABig{} ($D{=}1024$, 24~layers).}
Three variants are evaluated:
\emph{Base}, the raw pretrained model before any alignment fine-tuning;
\emph{Chat}, the instruction-tuned model;
and \emph{Abliterated}, the Chat model with the refusal direction removed from
all weight matrices via orthogonalisation~\citep{arditi2024refusal}, rendering
it unable to produce refusals.
Both non-base variants employ a hybrid linear-attention architecture; activations
were extracted using a standard PyTorch
implementation.\footnote{The \texttt{flash-linear-attention} fast path was
unavailable at the time of testing; results reflect geometric-signal robustness
across attention implementations.}

\paragraph{\QASmall{} ($D{=}896$, 24~layers).}
The same three variants (\emph{Base}, \emph{Instruct}, and \emph{Abliterated})
are evaluated for this family, with the abliterated variant produced by
orthogonalising the refusal direction from the Instruct checkpoint.

\subsection{Detection performance}
\label{sec:main_results}

\Cref{tab:main} summarises detection performance across all six variants.
Three findings hold without exception.
Harmful intent and benign-aggressive phrasing (XSTest) are perfectly separable
across all six models, including the abliterated variants: AUROC h/b $= 1.000$
universally.
The h/n task is strongly solved as well, with AUROC ranging from 0.9374
(\QASmall{}-Abliterated) to 0.9642 (\QABig{}-Base) and AUPRC h/n
$\geq$0.898 throughout.
Most importantly for our central claim, the abliterated models are essentially
indistinguishable from their instruction-tuned counterparts in detection
performance: the abliterated variant falls within 0.002 AUROC h/n of the
Chat model for \QABig{} and within 0.005 for \QASmall{}.
These margins are smaller than the gap between base and instruction-tuned
variants within either family. All pairwise comparisons achieve $p < 10^{-45}$ under the Mann--Whitney $U$
test (harmful vs.\ normative/rest); normative-vs-benign-agg $p$-values range
from $10^{-5}$ to $10^{-22}$.

\begin{table}[htb]
\centering
\caption{Detection performance (normative-reference strategy, $K{=}1$,
held-out evaluation sets, $N_\mathrm{fit}{=}200$ for all models).
$n_\mathrm{harm}{=}520$, $n_\mathrm{norm,eval}{=}520$,
$n_\mathrm{benign}{=}250$ for all models.
h/n: harmful vs.\ normative; h/b: harmful vs.\ benign-aggressive; h/r: harmful
vs.\ rest (norm $\cup$ benign).
$r_\mathrm{b/n}$: rank-biserial correlation for normative-vs-benign-agg
(negative means benign-agg is \emph{less} anomalous than normative, which is the
desired direction for a harm detector).
Prec@90: precision at 90\% recall on the h/n task.
$^\dagger$Best layer for this model is 10; AUROC at layer~20 differs by $<0.004$.
}
\label{tab:main}
\small
\setlength{\tabcolsep}{4.5pt}
\begin{tabular}{llrrrrrrrr}
\toprule
\multirow{2}{*}{Model} & \multirow{2}{*}{Type}
  & \multirow{2}{*}{Layer}
  & \multicolumn{2}{c}{AUROC} & \multicolumn{2}{c}{AUPRC}
  & \multirow{2}{*}{h/r} & \multirow{2}{*}{$r_\mathrm{b/n}$}
  & \multirow{2}{*}{Prec@90} \\
\cmidrule(lr){4-5}\cmidrule(lr){6-7}
 & & & h/n & h/b & h/n & h/b & & & \\
\midrule
\multicolumn{10}{l}{\textit{\QABig{}} ($D{=}1024$)} \\
Base        & Base        & 20 & 0.9642 & 1.000 & 0.9373 & 1.000 & 0.9758 & $-0.384^{***}$ & 0.928 \\
Chat        & Instruct    & 20 & 0.9497 & 1.000 & 0.9117 & 1.000 & 0.9661 & $-0.434^{***}$ & 0.899 \\
Abliterated & Abliterated & 20 & 0.9517 & 1.000 & 0.9165 & 1.000 & 0.9674 & $-0.427^{***}$ & 0.899 \\
\midrule
\multicolumn{10}{l}{\textit{\QASmall{}} ($D{=}896$)} \\
Base        & Base        & 20 & 0.9585 & 1.000 & 0.9373 & 1.000 & 0.9720 & $+0.149^{**}$  & 0.902 \\
Instruct    & Instruct    & 20 & 0.9420 & 1.000 & 0.9129 & 1.000 & 0.9609 & $+0.219^{***}$ & 0.875 \\
Abliterated$^\dagger$
            & Abliterated & 10 & 0.9374 & 1.000 & 0.8978 & 1.000 & 0.9577 & $-0.179^{**}$  & 0.882 \\
\bottomrule
\end{tabular}
\vspace{3pt}\\
\raggedright\footnotesize
$^{**}$: $p{<}10^{-4}$.\quad $^{***}$: $p{<}10^{-18}$.
All h/n, h/b, h/r comparisons: $p{<}10^{-45}$.
Benign-agg in \QASmall{}-Base and \QASmall{}-Instruct scores slightly above
normative ($r_\mathrm{b/n}{>}0$) but far below harmful (AUROC h/b $=1.000$).
\end{table}

\begin{figure}[htbp]
  \centering
  \begin{subfigure}[t]{0.48\textwidth}
    \includegraphics[width=\linewidth]{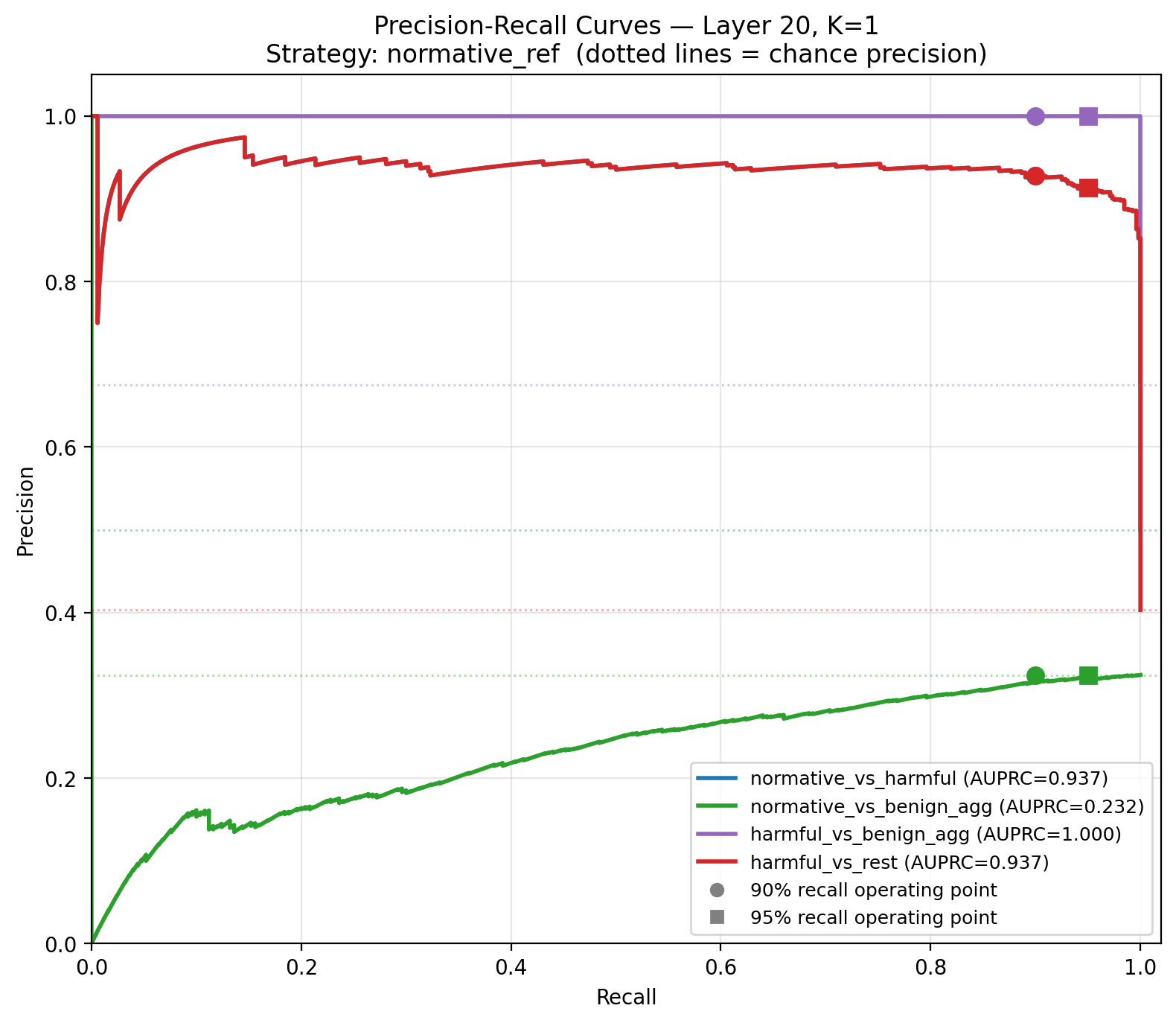}
    \caption{\QABig{}-Base}
  \end{subfigure}\hfill
  \begin{subfigure}[t]{0.48\textwidth}
    \includegraphics[width=\linewidth]{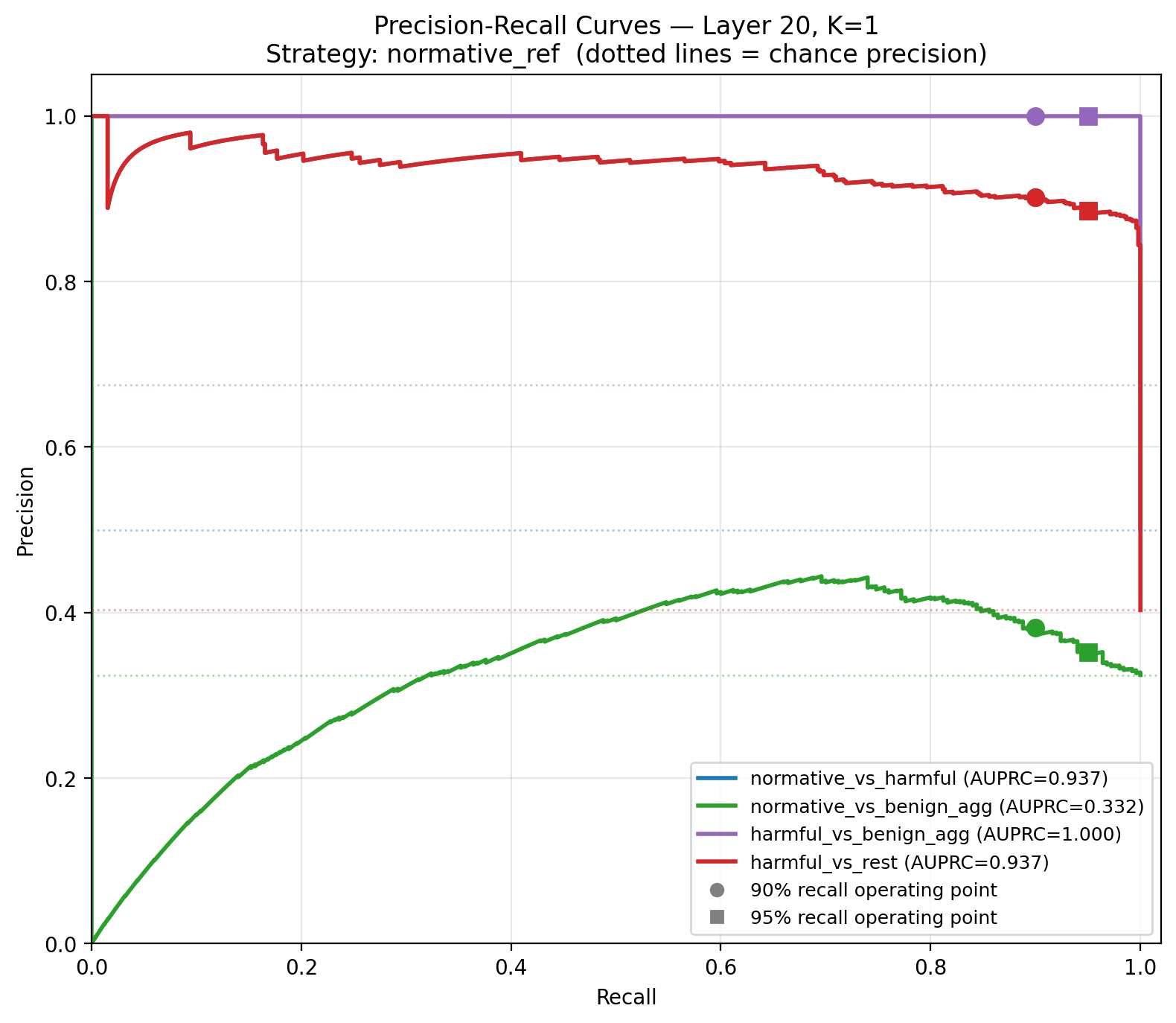}
    \caption{\QASmall{}-Base}
  \end{subfigure}
  \caption{%
    \textbf{Precision-recall curves at the operating layer ($K{=}1$)} for the
    two base variants; remaining models are in \cref{fig:app_pr_all}.
    Dotted horizontal lines indicate chance precision per task.
    The harmful-vs-normative curve (red) maintains precision $>0.92$ up to
    90\% recall for \QABig{}-Base (Prec@90 $= 0.928$) and $>0.90$ for
    \QASmall{}-Base (Prec@90 $= 0.902$).
    The harmful-vs-benign-agg curve is flat at 1.000 precision across all
    recall levels.
    The normative-vs-benign-agg curve (green) lies \emph{below} chance in
    \QABig{} (AUPRC $= 0.232$), confirming that benign-agg prompts are
    scored as less anomalous than normative, consistent with $r_\mathrm{b/n}
    = -0.384$.
  }
  \label{fig:pr_curve}
\end{figure}

Benign-aggressive placement is family-dependent but never a problem for
discrimination.
In \QABig{}, XSTest prompts score significantly \emph{below} normative
($r_\mathrm{b/n} \approx -0.43$ for Chat and Abliterated, $-0.38$ for Base),
occupying the innermost region of the normative ring.
In \QASmall{}-Base and Instruct, they sit slightly \emph{above} normative
($r_\mathrm{b/n} = +0.15$ and $+0.22$); in \QASmall{}-Abliterated, they
shift back below ($r_\mathrm{b/n} = -0.18$).
In every configuration, harmful prompts are far more anomalous than
benign-aggressive ones, achieving AUROC h/b $= 1.000$ throughout.
The full precision-recall profiles are shown in \cref{fig:pr_curve} for the
base variants and in \cref{fig:app_pr_all} for all six models.

\subsection{Comparison across model variants}
\label{sec:base_vs_instruct}

Within each family, base models match or slightly exceed their instruction-tuned
counterparts in harmful detection: the gap is 0.015 AUROC h/n for \QABig{}
and 0.017 for \QASmall{}, with base scoring higher in both cases.
This is a consistent if modest observation, and points to safety geometry being
present before alignment fine-tuning.

The abliterated models tell the more striking story.
In \QABig{}, the abliterated model (AUROC h/n $= 0.9517$) lies 0.002 above
the chat model ($0.9497$), well within any plausible noise floor.
In \QASmall{}, the gap is 0.005 (best layer 10 vs.\ 20), a margin smaller
than the base-to-instruct difference within the same family.
A model that cannot refuse harmful requests is, by this measure, an equally
capable harm detector.

\begin{figure}[htbp]
  \centering
  \begin{subfigure}[t]{\textwidth}
    \includegraphics[width=\linewidth]{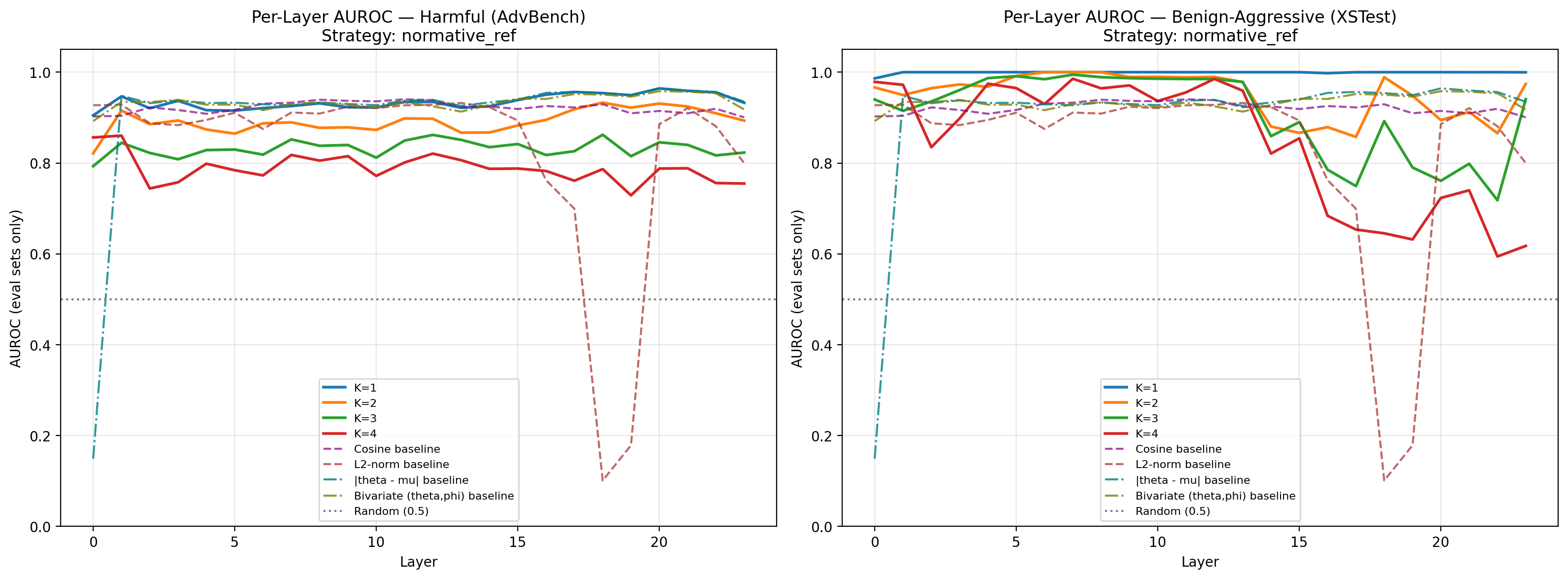}
    \caption{\QABig{}-Base}
  \end{subfigure}
  \\[6pt]
  \begin{subfigure}[t]{\textwidth}
    \includegraphics[width=\linewidth]{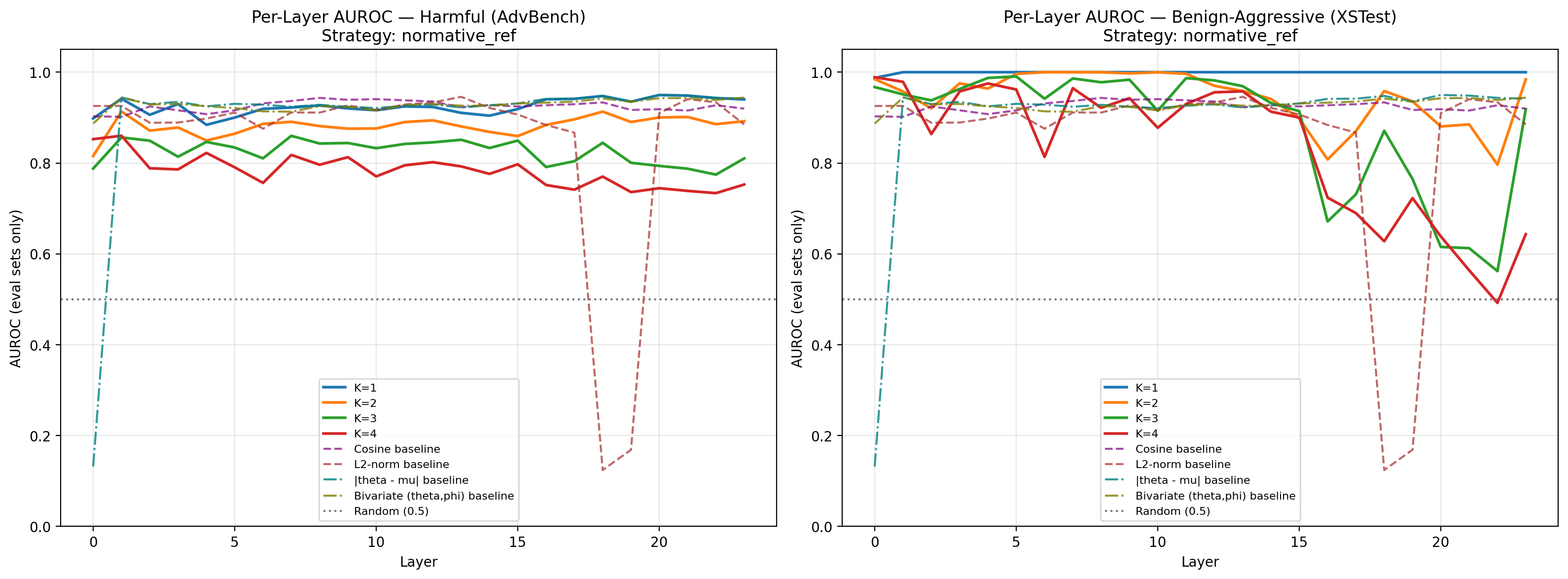}
    \caption{\QABig{}-Chat}
  \end{subfigure}
  \\[6pt]
  \begin{subfigure}[t]{\textwidth}
    \includegraphics[width=\linewidth]{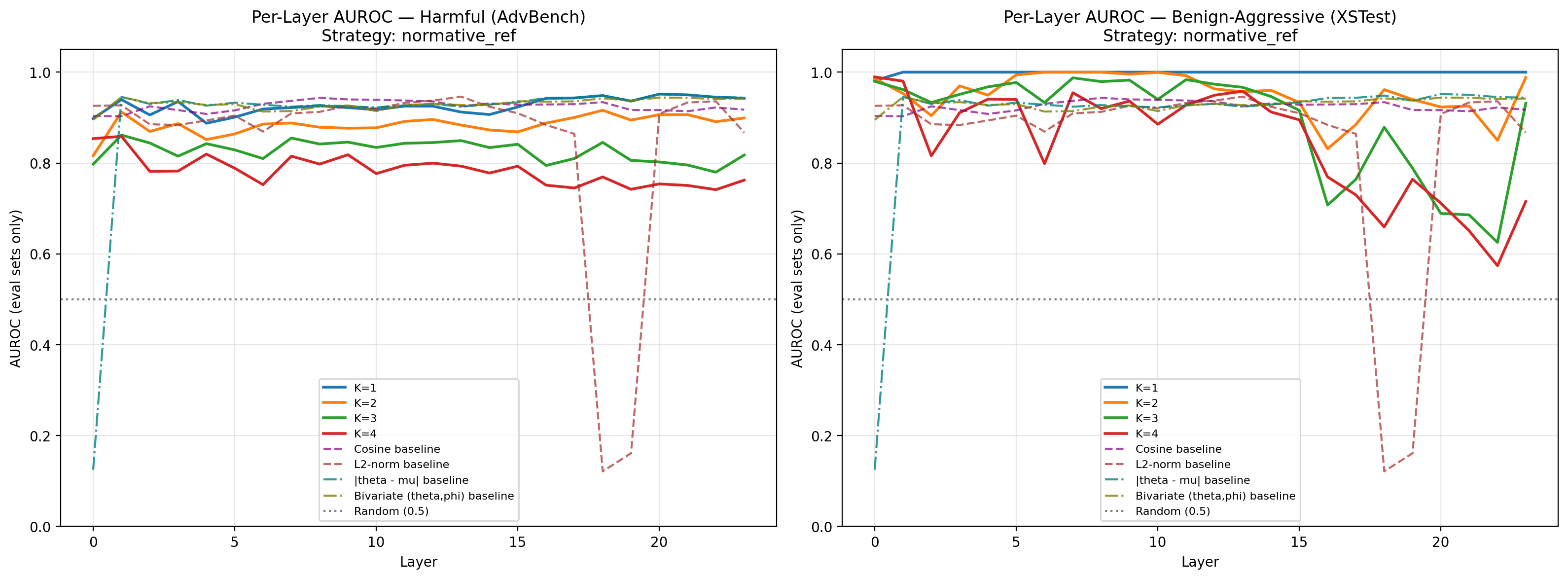}
    \caption{\QABig{}-Abliterated}
  \end{subfigure}
  \caption{%
    \textbf{Per-layer AUROC for the \QABig{} family.}
    Each panel shows AUROC h/n (left) and AUROC h/b (right) vs.\ layer index.
    $K{=}1$ (solid blue) strictly dominates $K{>}1$ (orange/green/red) at
    every layer.
    The cosine-centroid baseline (purple dashed) consistently underperforms
    $K{=}1$.
    AUROC h/b $= 1.000$ is maintained at every layer regardless of alignment
    stage. The three panels are nearly indistinguishable in their h/b profile.
  }
  \label{fig:auroc_by_layer_qwen35}
\end{figure}

\begin{figure}[htbp]
  \centering
  \begin{subfigure}[t]{\textwidth}
    \includegraphics[width=\linewidth]{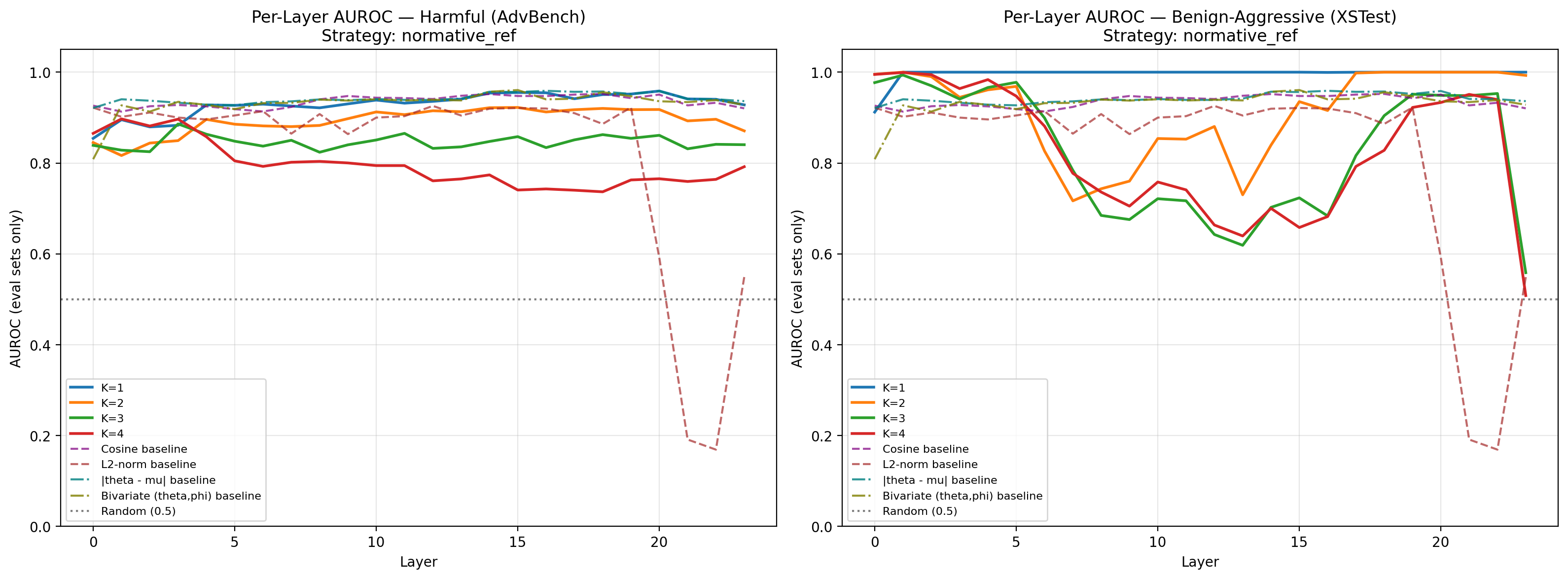}
    \caption{\QASmall{}-Base}
  \end{subfigure}
  \\[6pt]
  \begin{subfigure}[t]{\textwidth}
    \includegraphics[width=\linewidth]{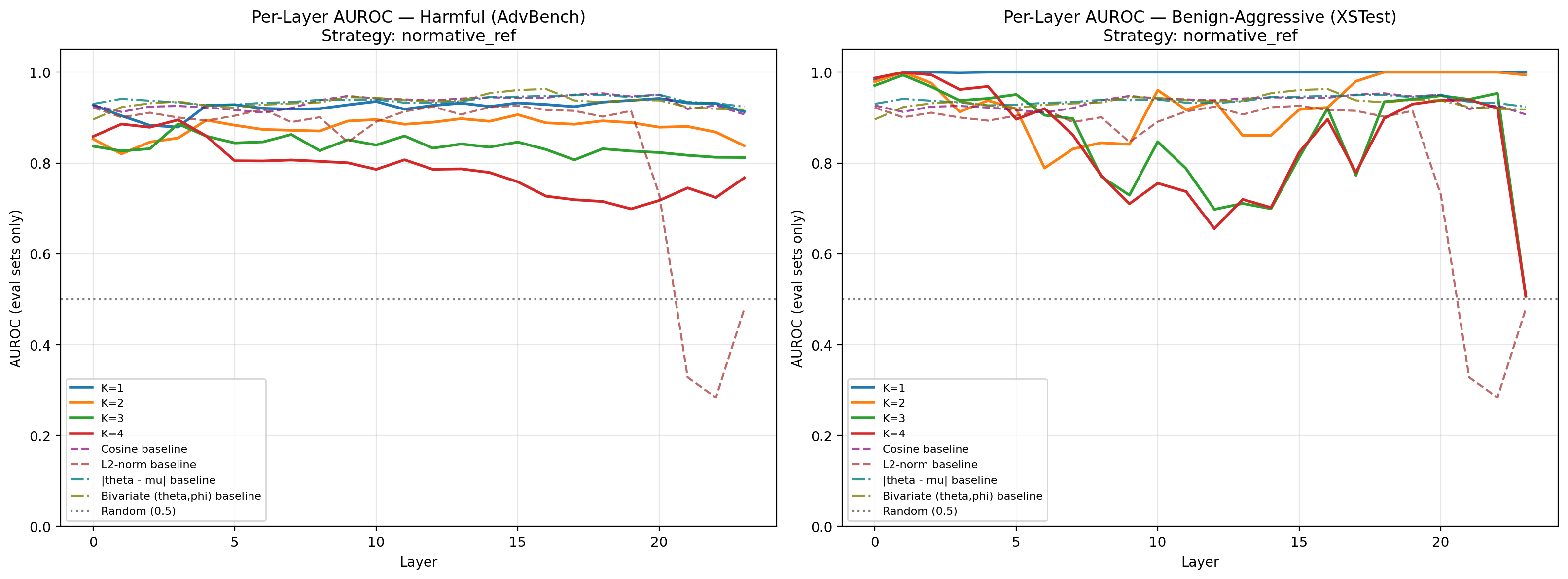}
    \caption{\QASmall{}-Instruct}
  \end{subfigure}
  \\[6pt]
  \begin{subfigure}[t]{\textwidth}
    \includegraphics[width=\linewidth]{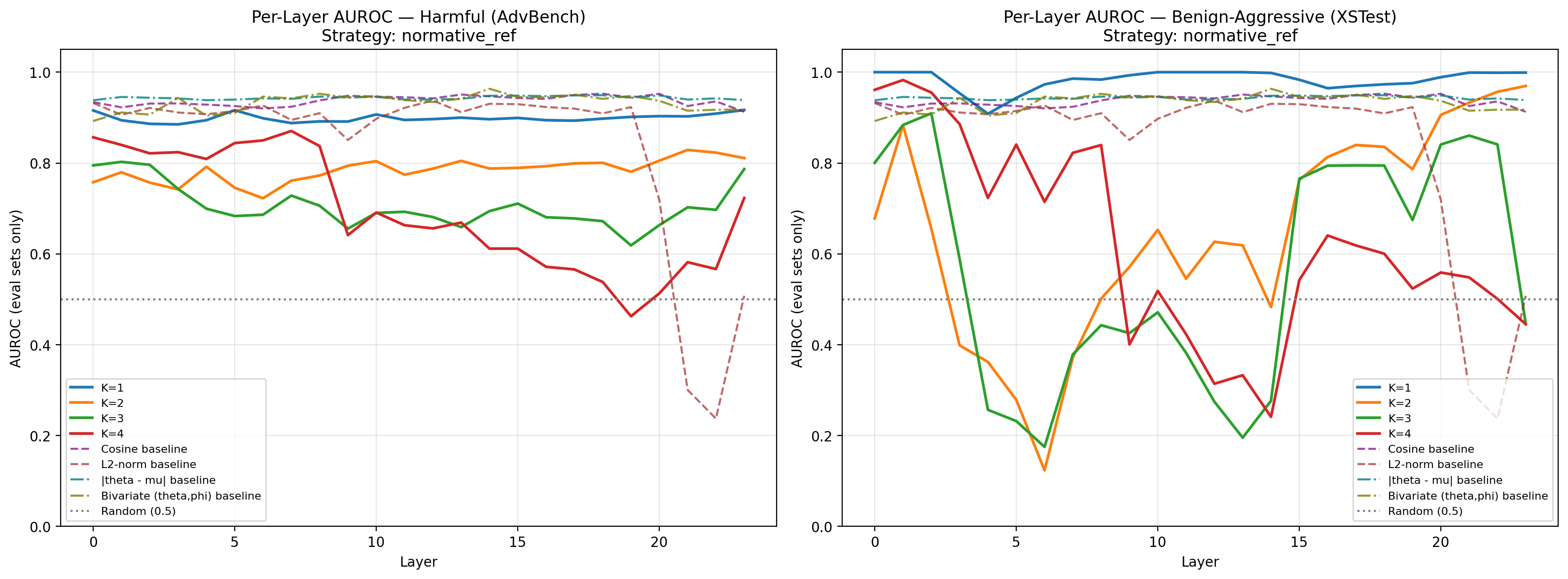}
    \caption{\QASmall{}-Abliterated}
  \end{subfigure}
  \caption{%
    \textbf{Per-layer AUROC for the \QASmall{} family.}
    $K{=}1$ (solid blue) remains the dominant scorer throughout.
    In the Instruct and Abliterated variants, the cosine-centroid and L2-norm
    baselines reach near-parity with $K{=}1$, reflecting the specific geometry
    of this family, but do not surpass it.
    The broad performance plateau across layers 5--23 bounds layer-selection
    optimism to $<0.08$ AUROC units.
  }
  \label{fig:auroc_by_layer_qwen25}
\end{figure}

\paragraph{Per-layer profile.}
\Cref{fig:auroc_by_layer_qwen35,fig:auroc_by_layer_qwen25} show per-layer
AUROC for both families.
$K{=}1$ is uniformly superior to $K{>}1$ baselines across all 24 layers in
every variant.
The broad plateau at $K=1$ bounds layer-selection optimism tightly: best layer exceeds worst by fewer than 0.08 AUROC units,
with all layers above 0.88.

\subsection{Computational cost}
\label{sec:cost}

We evaluated the latency overhead of \method{} on an NVIDIA GeForce
RTX~3070 Laptop GPU (8\,GB VRAM), reporting mean and standard deviation
over 100 trials after GPU warm-up, using Alpaca prompts to reflect realistic
sequence lengths. We report mean (standard deviation):
A baseline \QASmall{} forward pass takes 20.7~(2.7)\,ms; residual-stream
extraction adds negligible overhead (0.0\,ms); and anomaly scoring (dot product
plus scalar Gaussian NLL) executes in 0.43~(0.08)\,ms.
End-to-end, the pipeline completes in 22.6~(2.1)\,ms per query.
Normative reference fitting is performed once offline: for $N{=}200$ prompts,
activation extraction takes $<$3\,s and PCA completes in $<$0.2\,s on CPU.
\method{} therefore adds less than 0.5\,ms of method-specific overhead
(anomaly scoring) relative to a standard forward pass.

\section{Geometric Analysis: The Two-Ring Structure}
\label{sec:geometry}
\begin{figure}[htbp]
  \centering
  \footnotesize 
  
  \begin{subfigure}[t]{0.44\textwidth}
    \includegraphics[width=\linewidth]{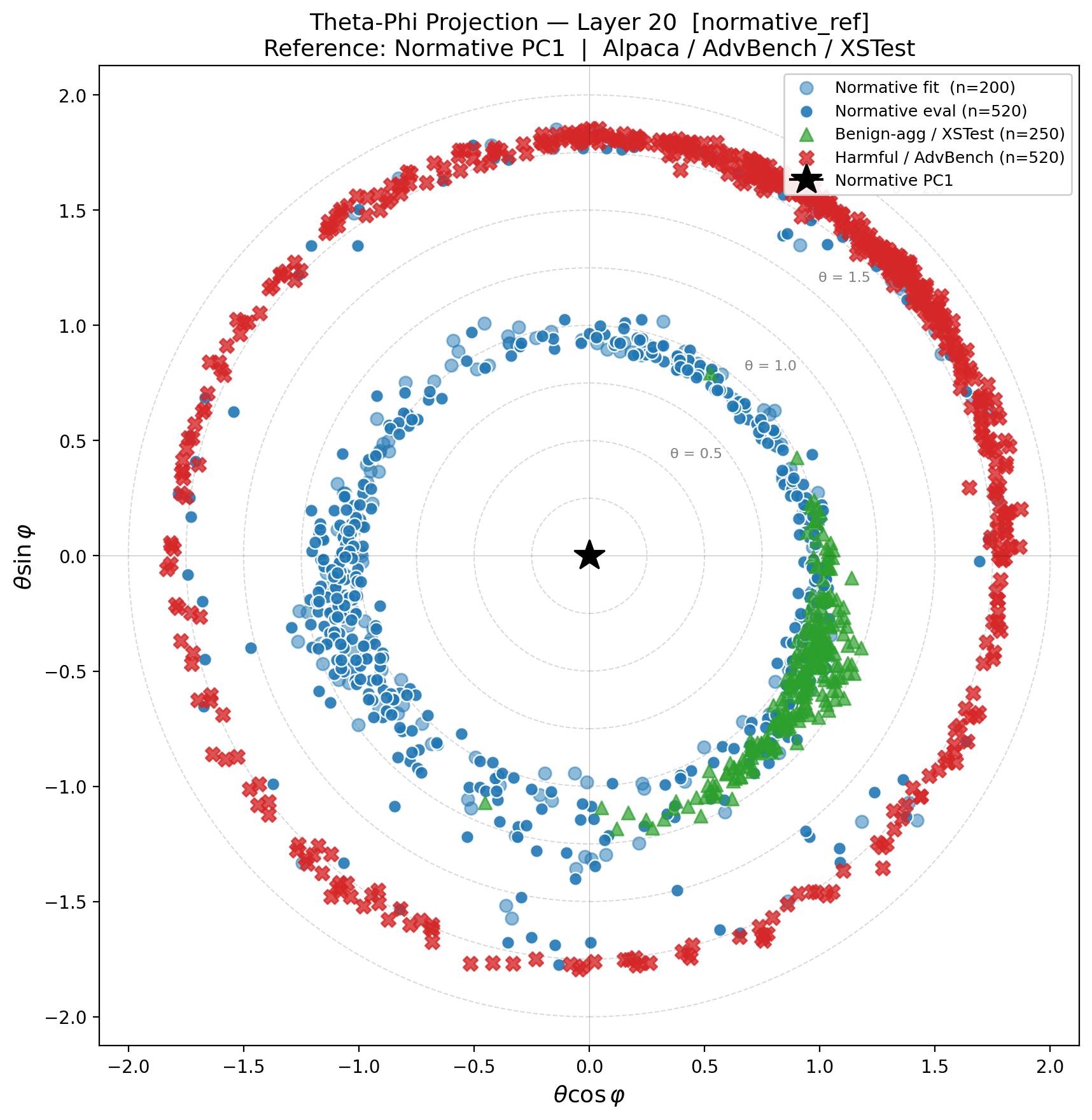}
    \caption{\QABig{}-Base (L20)}
  \end{subfigure}\hfill
  \begin{subfigure}[t]{0.44\textwidth}
    \includegraphics[width=\linewidth]{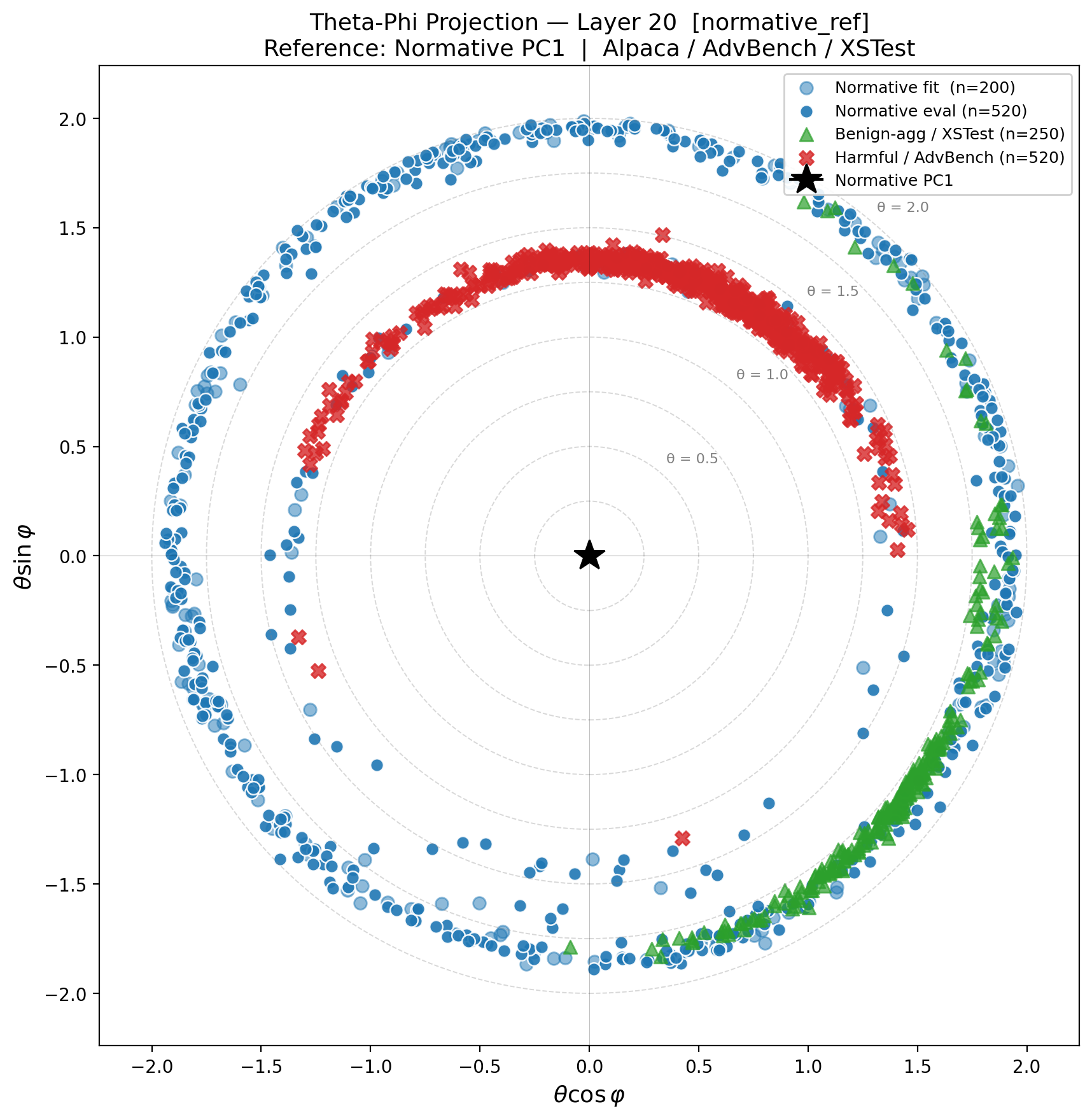}
    \caption{\QASmall{}-Base (L20)}
  \end{subfigure}
  \\[2pt] 
  
  \begin{subfigure}[t]{0.44\textwidth}
    \includegraphics[width=\linewidth]{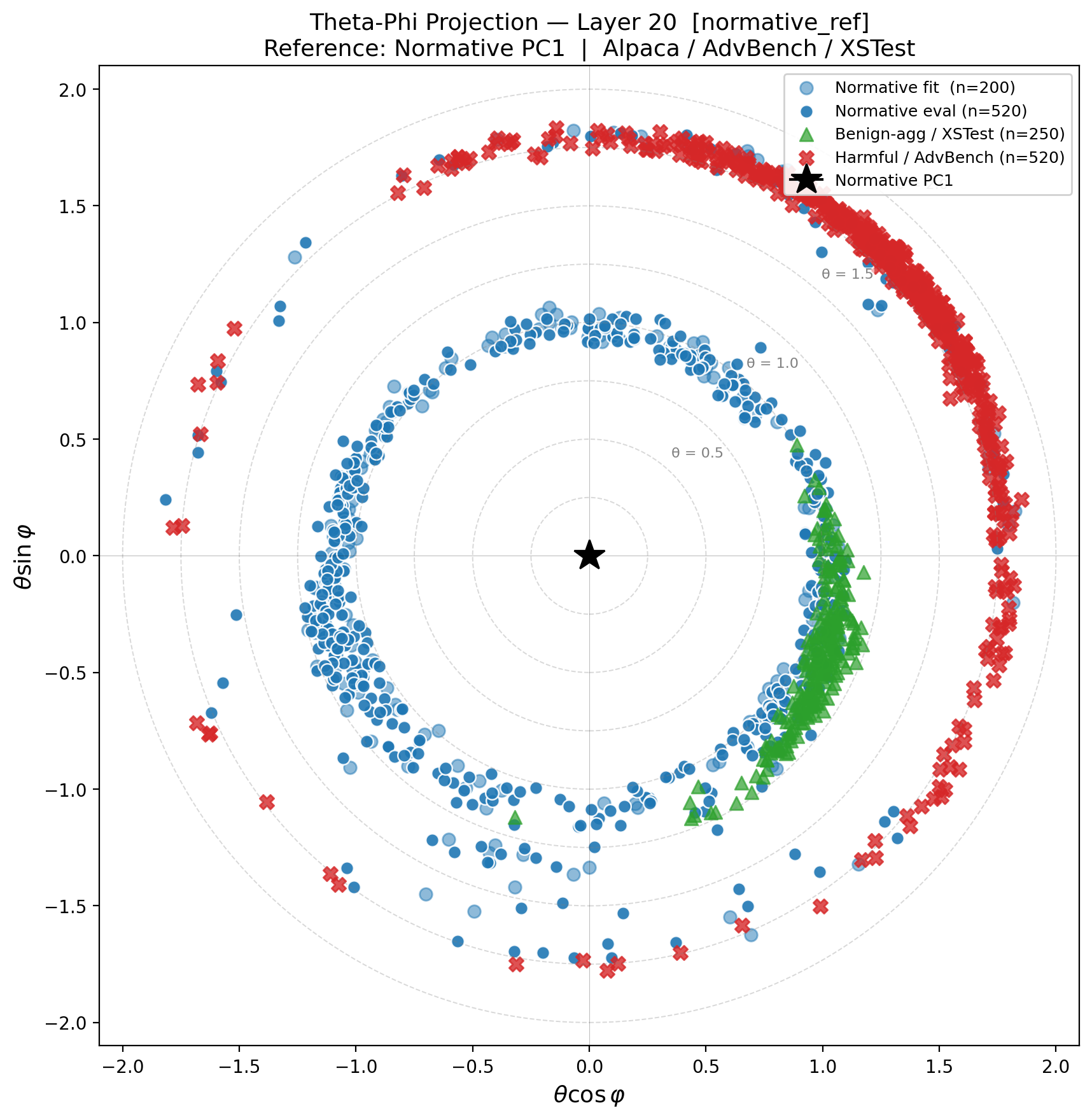}
    \caption{\QABig{}-Chat (L20)}
  \end{subfigure}\hfill
  \begin{subfigure}[t]{0.44\textwidth}
    \includegraphics[width=\linewidth]{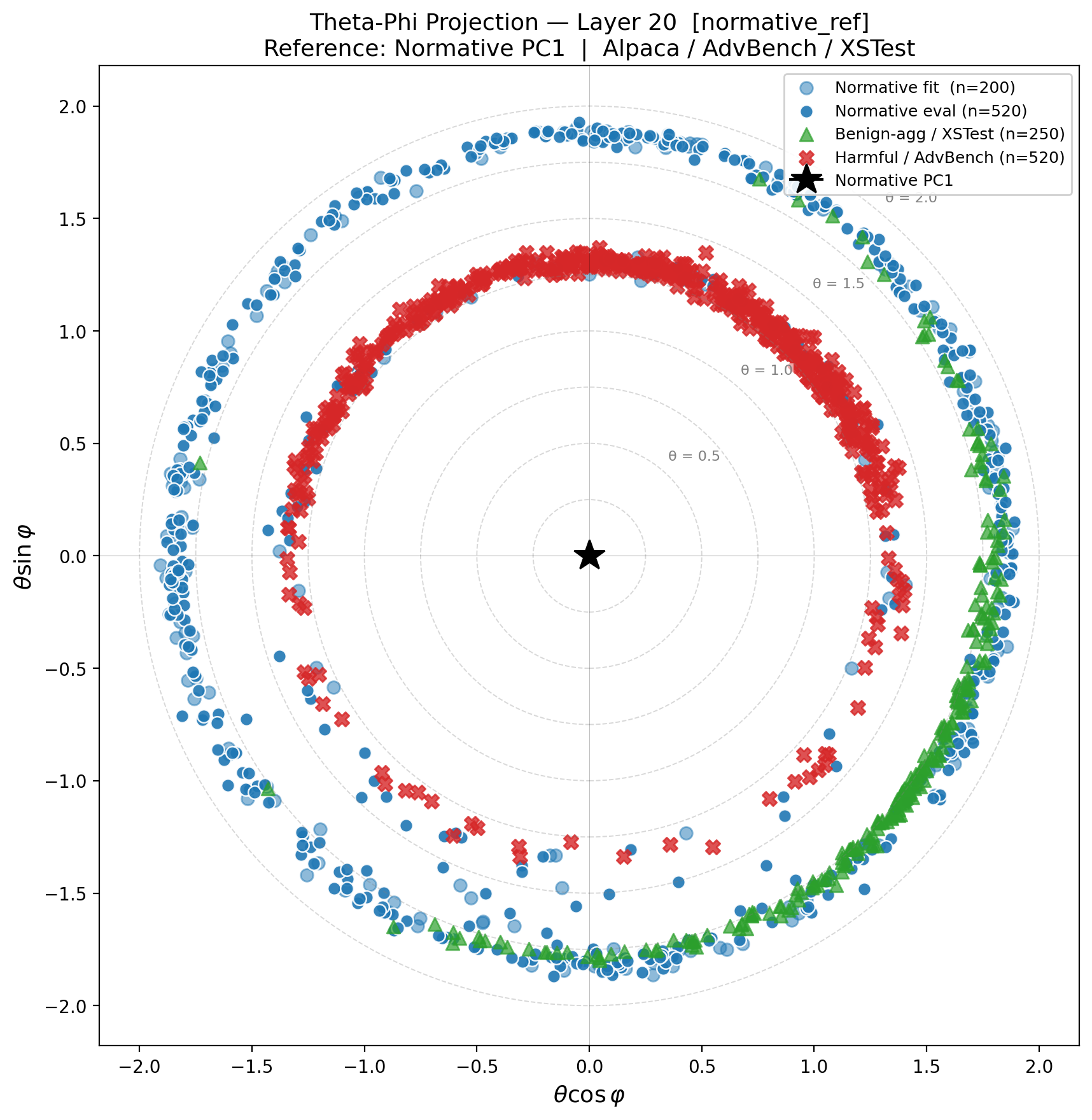}
    \caption{\QASmall{}-Instruct (L20)}
  \end{subfigure}
  \\[2pt]
  
  \begin{subfigure}[t]{0.44\textwidth}
    \includegraphics[width=\linewidth]{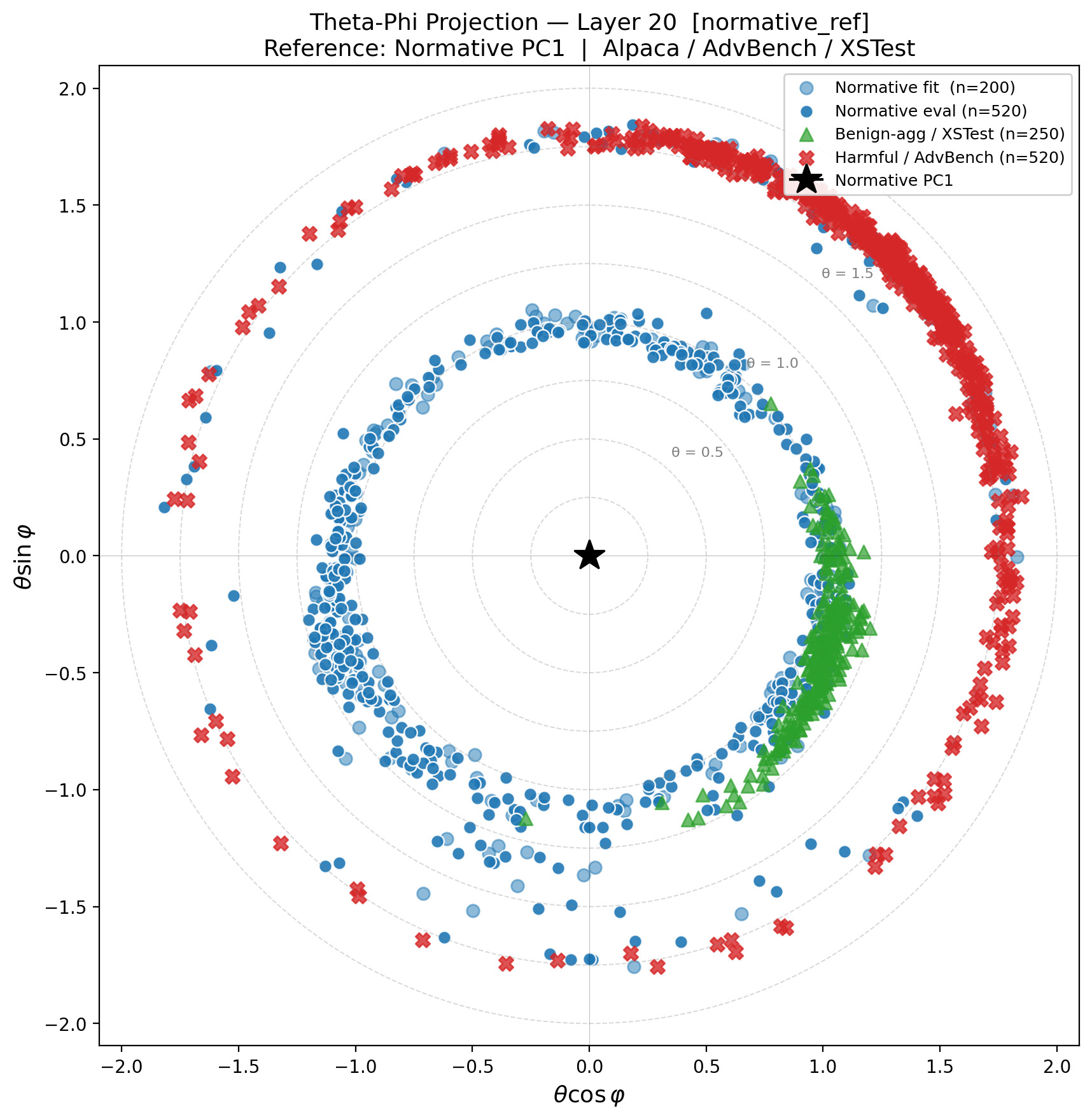}
    \caption{\QABig{}-Abliterated (L20)}
  \end{subfigure}\hfill
  \begin{subfigure}[t]{0.44\textwidth}
    \includegraphics[width=\linewidth]{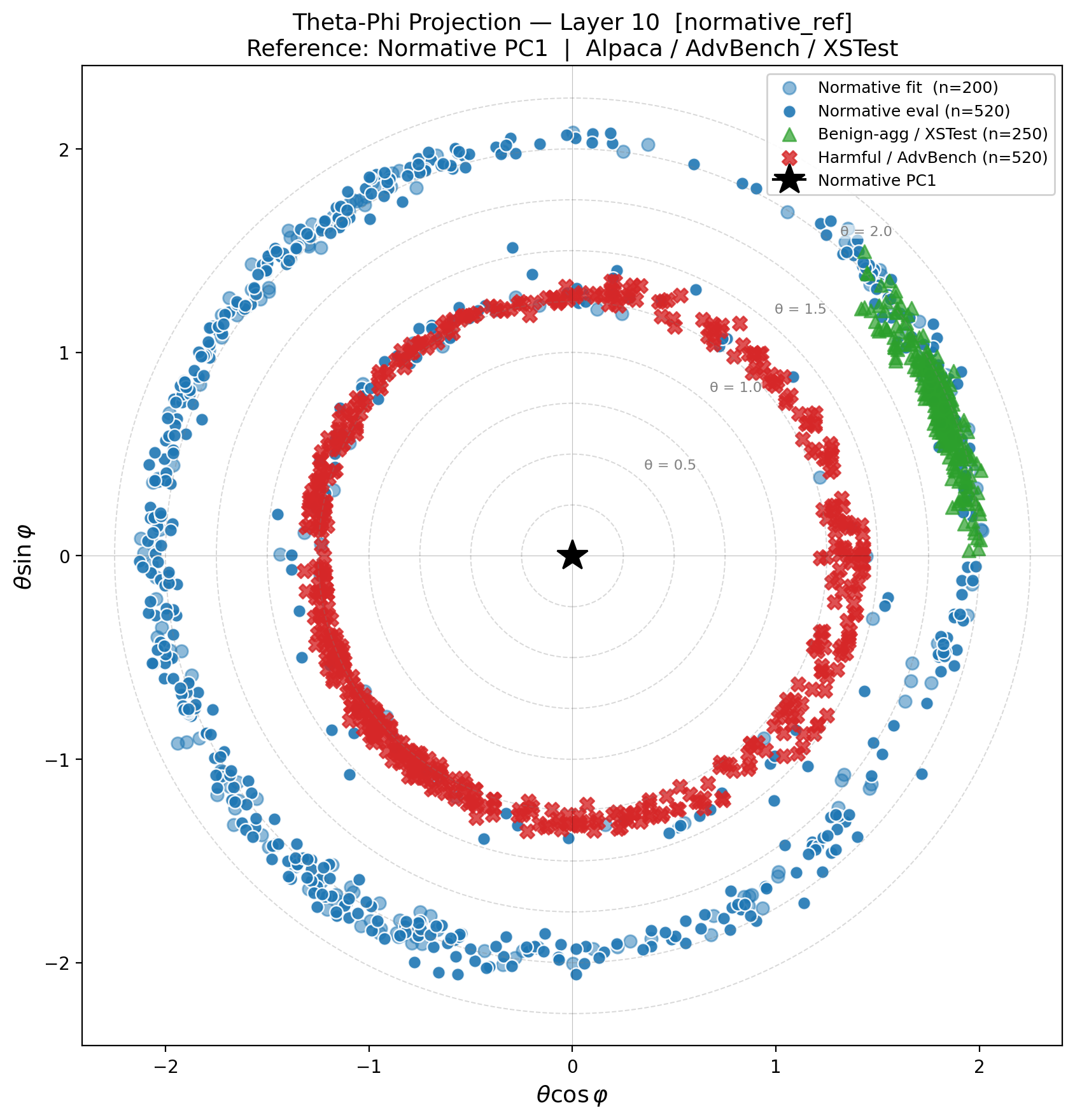}
    \caption{\QASmall{}-Abliterated (L10)}
  \end{subfigure}
  
  \vspace{-5pt} 
  \caption{%
    \small \textbf{Theta-phi projections at the operating layer.}
    Harmful (red) and safe (blue/green) prompts form distinct concentric radial zones across all variants. 
    In the \QABig{} family (\textbf{left column}), harmful intent occupies the \textbf{outer ring}; in \QASmall{} (\textbf{right column}), it occupies the \textbf{inner ring}. 
    The visual invariance across rows demonstrates that safety geometry is established during pretraining and remains intact even after the mathematical erasure of refusal mechanisms. All panels achieve AUROC h/b = 1.000.
  }
  \label{fig:theta_phi}
\end{figure}

\subsection{Opposite ring orientations across families}
\label{sec:ring_direction}

\Cref{fig:theta_phi} plots the theta-phi projection for all six variants at
their respective operating layers.
A concentric two-ring structure separating harmful and normative activations is
present in every panel, and the two families occupy \emph{opposite} positions.

In all three \QABig{} variants, harmful prompts occupy the \emph{outer} ring
($\bar\theta_\mathrm{harm} \approx 1.80$~rad vs.\ $\mu_0 \approx 1.17$~rad;
\cref{tab:theta_stats}): they deviate more strongly from PC1 than normative
prompts.
In all three \QASmall{} variants, harmful prompts occupy the \emph{inner} ring
($\bar\theta_\mathrm{harm} \approx 1.32$--$1.36$~rad vs.\
$\mu_0 \approx 1.76$--$1.90$~rad): they are more tightly aligned with PC1
than normative prompts.

This family-level reversal cannot be reconciled by any fixed directional
threshold on $\theta$.
The anomaly score $s(x)$ resolves it correctly in all cases: both
$\theta = \mu_0 + k\sigma_0$ and $\theta = \mu_0 - k\sigma_0$ receive the
same score for any $k$, regardless of ring direction.
The same argument applies to layer-level reversals within each model, visible
in \cref{fig:auroc_by_layer_qwen35,fig:auroc_by_layer_qwen25} through the
changing sign of the L2-norm baseline across layers.

\subsection{Benign-aggressive placement}

The XSTest prompts occupy a consistent position relative to the harmful cluster
in each family, reinforcing the interpretation of the two-ring geometry.
In all three \QABig{} variants, they cluster at the smallest radii, below the
normative mean ($r_\mathrm{b/n} \approx -0.43$), and are geometrically
separated from harmful prompts in the opposite direction.
In \QASmall{}-Base and Instruct, they sit near or slightly above normative
($r_\mathrm{b/n} = +0.15$, $+0.22$), while in \QASmall{}-Abliterated they
shift slightly below ($r_\mathrm{b/n} = -0.18$).
In every case, perfect harmful-vs-benign-aggressive separation (AUROC h/b
$= 1.000$) holds, confirming that the theta score cleanly discriminates harmful
intent from aggressive but benign phrasing regardless of where benign-aggressive
prompts sit relative to the normative mean.

\subsection{Angular deviation statistics}
\label{sec:theta_stats}

\begin{figure}[htbp]
  \centering
  
  \begin{subfigure}[t]{0.48\textwidth}
    \includegraphics[width=\linewidth]{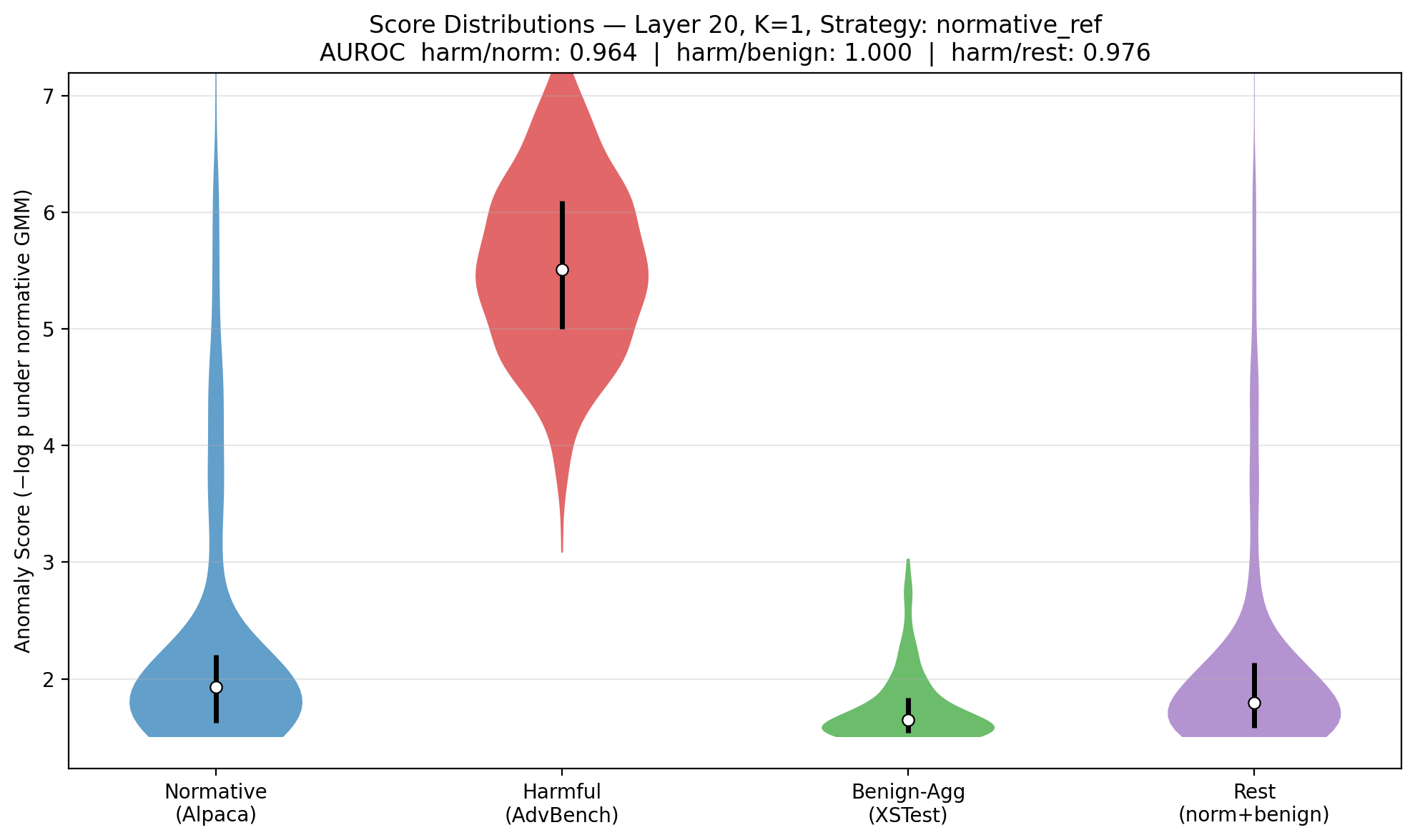}
    \caption{\QABig{}-Base}
  \end{subfigure}\hfill
  \begin{subfigure}[t]{0.48\textwidth}
    \includegraphics[width=\linewidth]{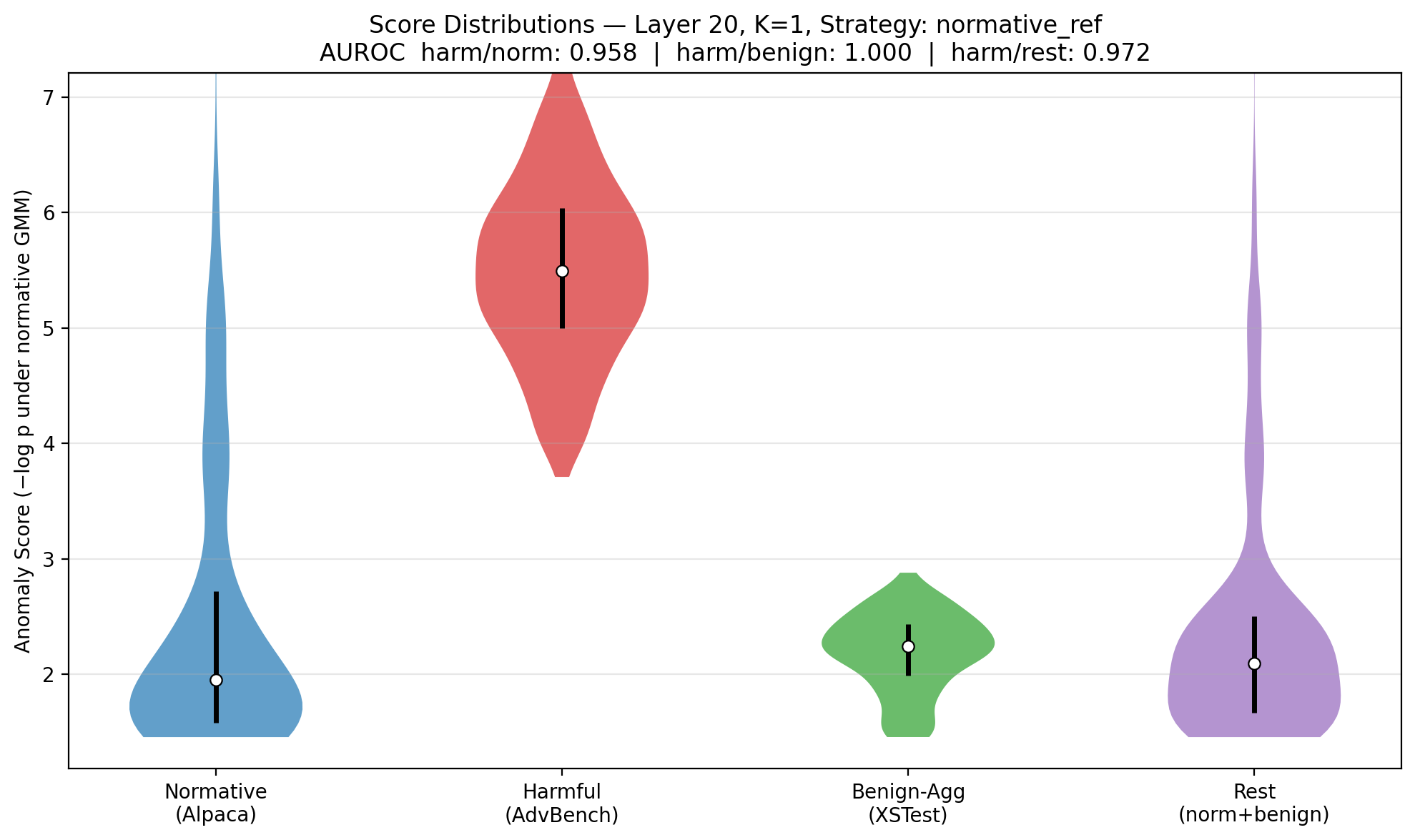}
    \caption{\QASmall{}-Base}
  \end{subfigure}
  \\[8pt]
  
  \begin{subfigure}[t]{0.48\textwidth}
    \includegraphics[width=\linewidth]{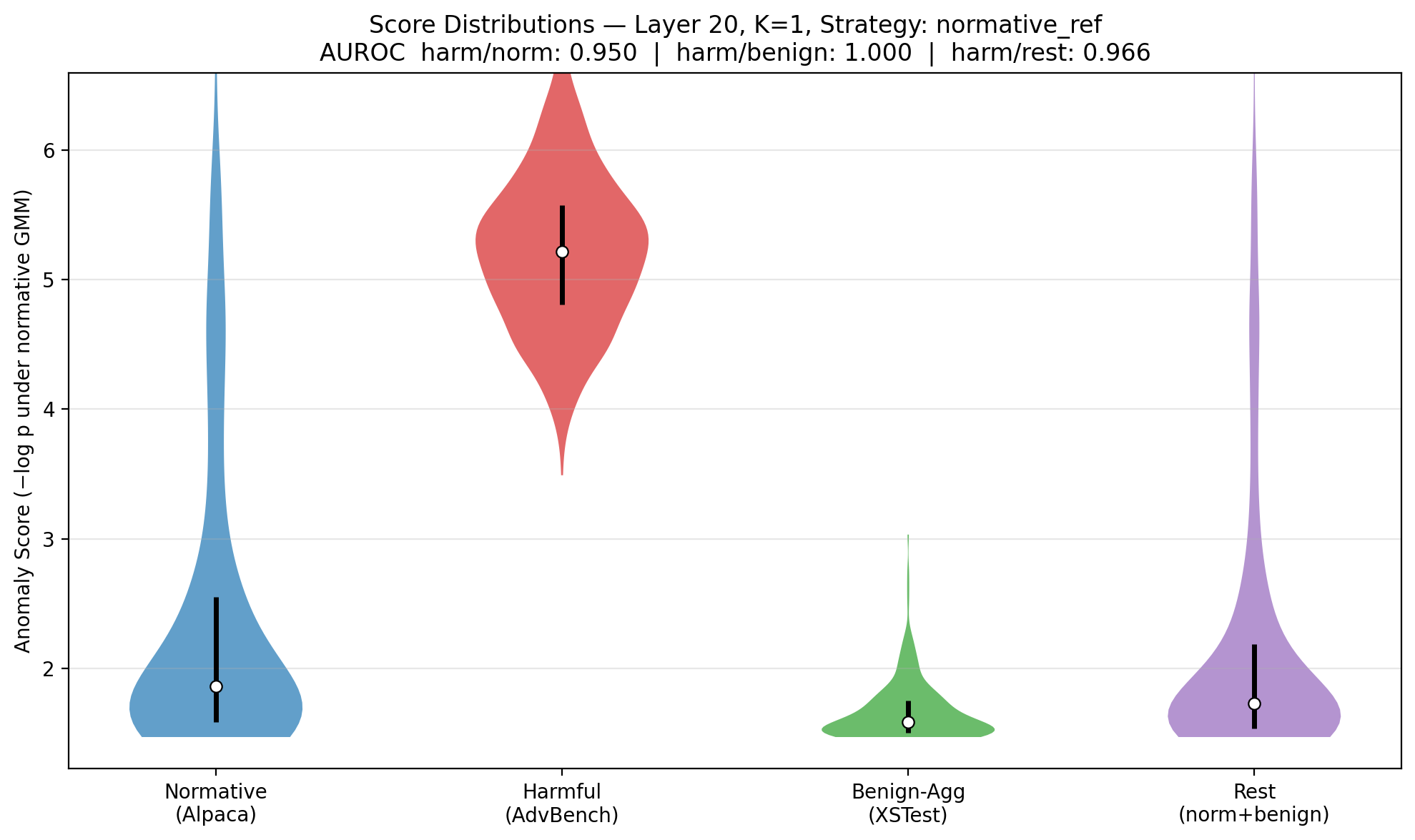}
    \caption{\QABig{}-Chat}
  \end{subfigure}\hfill
  \begin{subfigure}[t]{0.48\textwidth}
    \includegraphics[width=\linewidth]{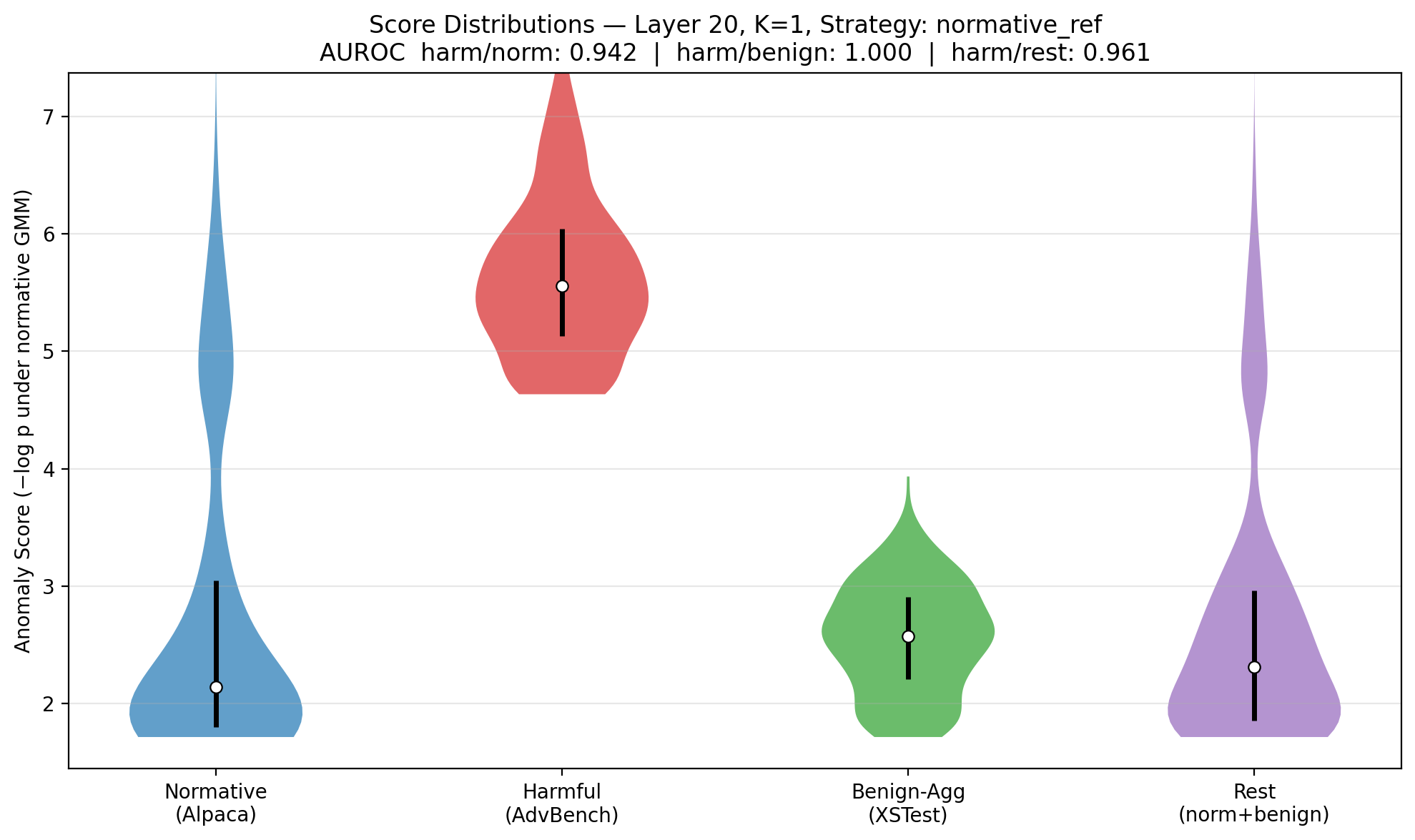}
    \caption{\QASmall{}-Instruct}
  \end{subfigure}
  \\[8pt]
  
  \begin{subfigure}[t]{0.48\textwidth}
    \includegraphics[width=\linewidth]{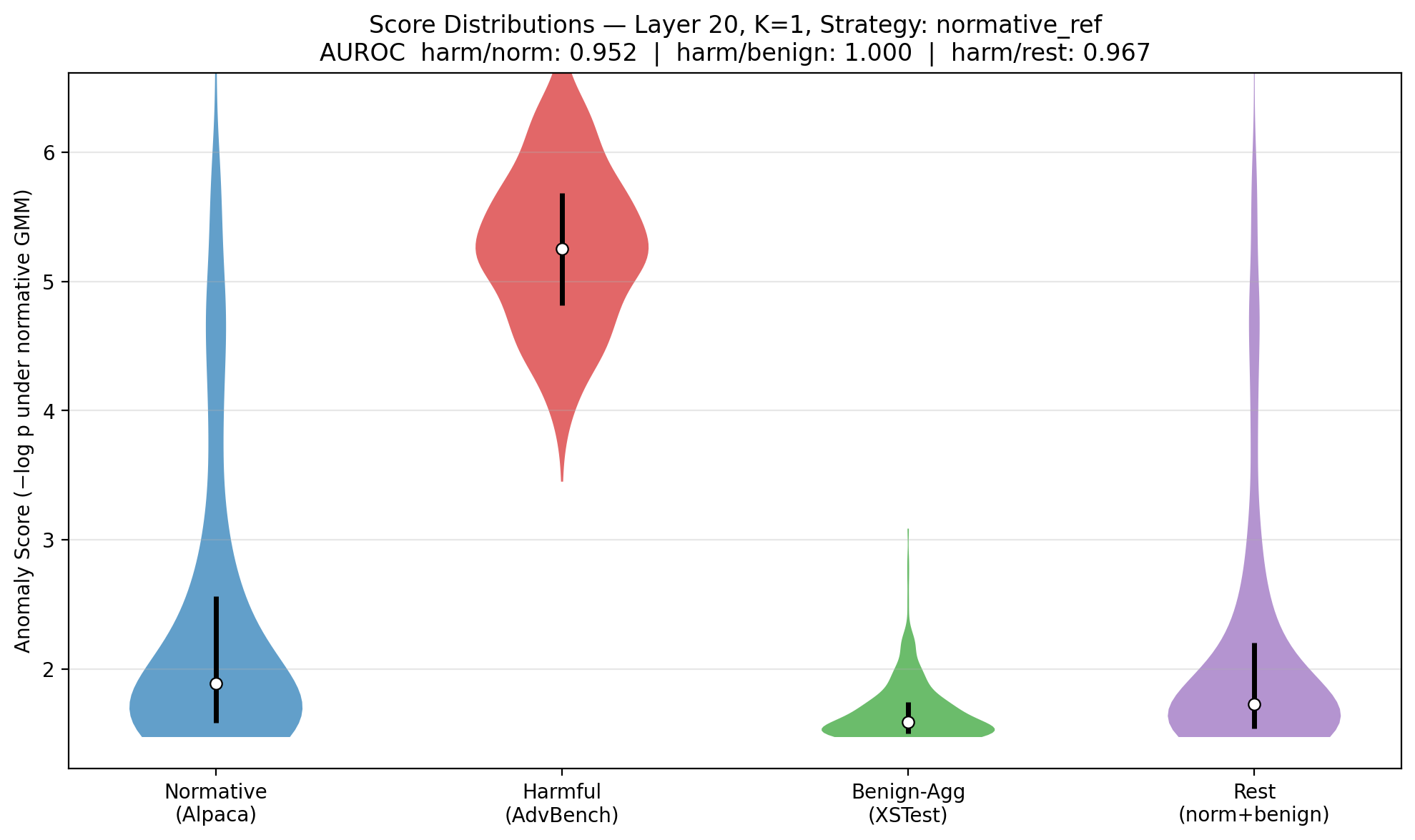}
    \caption{\QABig{}-Abliterated}
  \end{subfigure}\hfill
  \begin{subfigure}[t]{0.48\textwidth}
    \includegraphics[width=\linewidth]{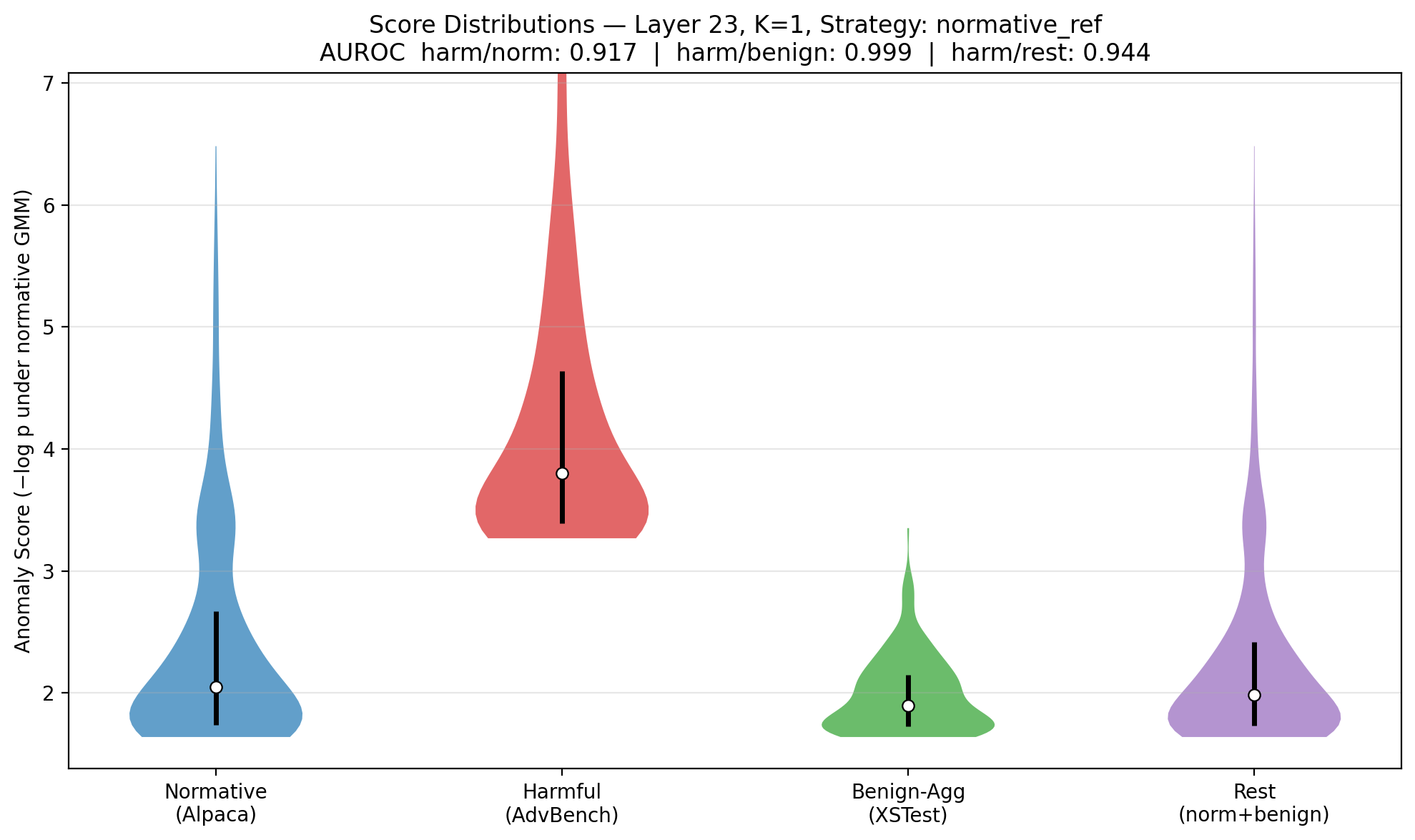}
    \caption{\QASmall{}-Abliterated}
  \end{subfigure}
  
  \caption{%
    \textbf{Anomaly score distributions at the operating layer for all six
    variants.}
    Each violin shows the marginal distribution of $s(x) = -\log p(\theta \mid \mu_0, \sigma_0^2)$ 
    for normative eval (blue), harmful (red), benign-aggressive (green), and 
    normative $\cup$ benign (purple). White circles denote medians; bars denote IQRs.
    In every panel, harmful prompts occupy a narrow, elevated band ($\sigma_\theta^\mathrm{harm}$
    is 5--9$\times$ smaller than $\sigma_\theta^\mathrm{norm}$; \cref{tab:theta_stats}).
    In the \QABig{} family (left column), benign-aggressive scores fall \emph{below} the
    normative distribution; in the \QASmall{} family (right column), they overlap with it.
    The three panels within each column are nearly identical, illustrating that
    abliteration leaves the score landscape intact.
  }
  \label{fig:score_dist}
\end{figure}

\Cref{tab:theta_stats} reports the raw angular deviation statistics at the
operating layer for each model, and \cref{fig:score_dist} shows the
corresponding anomaly score distributions.
Three patterns hold universally.

The harmful cluster is extraordinarily compact.
$\sigma_\theta^\mathrm{harm}$ is 0.030--0.052~rad: one order of magnitude
smaller than $\sigma_\theta^\mathrm{norm}$ (0.183--0.272~rad), and this
near-degeneracy is preserved across base, instruction-tuned, and abliterated
variants alike.
Normative train and test are statistically indistinguishable
($|\bar\theta_\mathrm{norm,train} - \bar\theta_\mathrm{norm,test}| \leq 0.01$~rad
in every model), confirming that $N{=}200$ defines a stable reference
distribution.
Finally, benign-aggressive prompts cluster near the normative mean in all
\QABig{} variants (smaller $\theta$ than normative, tighter alignment with
PC1), while in \QASmall{} their $\theta$ nearly coincides with the normative
$\mu_0$, leaving harmful as the angular outlier in both cases.

\begin{table}[htb]
\centering
\caption{Raw angular deviation $\theta$ statistics at the operating layer
(normative-ref, $K{=}1$).
$\mu_0 = \bar\theta_\mathrm{norm,test}$: mean normative angle (radians).
$\sigma_\mathrm{norm}$: normative test standard deviation.
$\bar\theta_\mathrm{harm}$: harmful mean.
$\Delta\theta = \bar\theta_\mathrm{harm} - \mu_0$: signed angular separation
(positive $=$ outer ring, negative $=$ inner ring).
$\bar\theta_\mathrm{benign}$: benign-aggressive mean.
$\sigma_\mathrm{harm}$: harmful standard deviation.
All h/n comparisons: $p{<}10^{-45}$.
$^\dagger$Layer 10 for this model.}
\label{tab:theta_stats}
\small
\begin{tabular}{llrrrrrrr}
\toprule
Model & Type & $\mu_0$ & $\sigma_\mathrm{norm}$ & $\bar\theta_\mathrm{harm}$
  & $\Delta\theta$ & $\bar\theta_\mathrm{benign}$ & $\sigma_\mathrm{harm}$ \\
\midrule
\multicolumn{8}{l}{\textit{\QABig{}}} \\
Base        & Base        & 1.161 & 0.272 & 1.811 & $+0.650$ & 1.094 & 0.034 \\
Chat        & Instruct    & 1.178 & 0.267 & 1.801 & $+0.623$ & 1.104 & 0.030 \\
Abliterated & Abliterated & 1.175 & 0.267 & 1.802 & $+0.627$ & 1.104 & 0.031 \\
\midrule
\multicolumn{8}{l}{\textit{\QASmall{}}} \\
Base        & Base        & 1.819 & 0.188 & 1.357 & $-0.462$ & 1.821 & 0.034 \\
Instruct    & Instruct    & 1.764 & 0.183 & 1.316 & $-0.448$ & 1.777 & 0.035 \\
Abliterated$^\dagger$
            & Abliterated & 1.904 & 0.250 & 1.301 & $-0.603$ & 1.962 & 0.052 \\
\bottomrule
\end{tabular}
\vspace{3pt}\\
\raggedright\footnotesize
$\Delta\theta>0$: harmful is outer ring (more deviated from PC1);
$\Delta\theta<0$: harmful is inner ring (more aligned with PC1).
Both configurations are correctly flagged by the symmetric score $s(x)$.
\end{table}

\section{Robustness and Sensitivity Analysis}
\label{sec:robustness}

\subsection{Safety signal dimensionality ($K$-ablation)}
\label{sec:ablation_k}

The per-layer AUROC plots already show that $K{=}1$ strictly outperforms
multi-directional baselines at every layer and model.
\Cref{tab:kablation} quantifies this at the operating layer.
Increasing from $K{=}1$ to the best $K{>}1$ reduces AUROC by 0.033--0.063
across models, a consistent and meaningful penalty for adding directions.
The cosine-centroid baseline matches or slightly exceeds $K{=}1$ for the two
\QASmall{} non-abliterated variants ($\Delta_\mathrm{cos} = +0.009$ and
$+0.004$), suggesting that for this family the mean normative activation is a
good proxy for PC1.
The PC1 formulation is nonetheless preferred on theoretical grounds: it
maximises captured normative variance, remains interpretable through the
theta-phi projection, and performs at least as well as the centroid in four of
six models.

\begin{table}[htb]
\centering
\caption{$K$-ablation at the operating layer.
$\Delta_K$: AUROC h/n change from $K{=}1$ to best $K{>}1$ (negative = $K{=}1$
better).
$\Delta_\mathrm{cos}$: AUROC h/n change from $K{=}1$ to cosine-centroid
baseline (negative = $K{=}1$ better).
$^\dagger$Layer 10.}
\label{tab:kablation}
\small
\begin{tabular}{llrrr}
\toprule
Model & Type & $K{=}1$ AUROC & $\Delta_K$ & $\Delta_\mathrm{cos}$ \\
\midrule
\QABig{}-Base        & Base        & 0.9642 & $-0.033$ & $-0.050$ \\
\QABig{}-Chat        & Instruct    & 0.9497 & $-0.049$ & $-0.031$ \\
\QABig{}-Abliterated & Abliterated & 0.9517 & $-0.045$ & $-0.035$ \\
\QASmall{}-Base      & Base        & 0.9585 & $-0.041$ & $-0.008$ \\
\QASmall{}-Instruct  & Instruct    & 0.9420 & $-0.063$ & $+0.009$ \\
\QASmall{}-Abliterated$^\dagger$
                     & Abliterated & 0.9374 & $-0.037$ & $+0.004$ \\
\bottomrule
\end{tabular}
\end{table}

\subsection{Safety signal sparsity (dimension ablation)}
\label{sec:ablation_dim}

\begin{figure}[htbp]
  \centering
  \begin{subfigure}[t]{0.48\textwidth}
    \includegraphics[width=\linewidth]{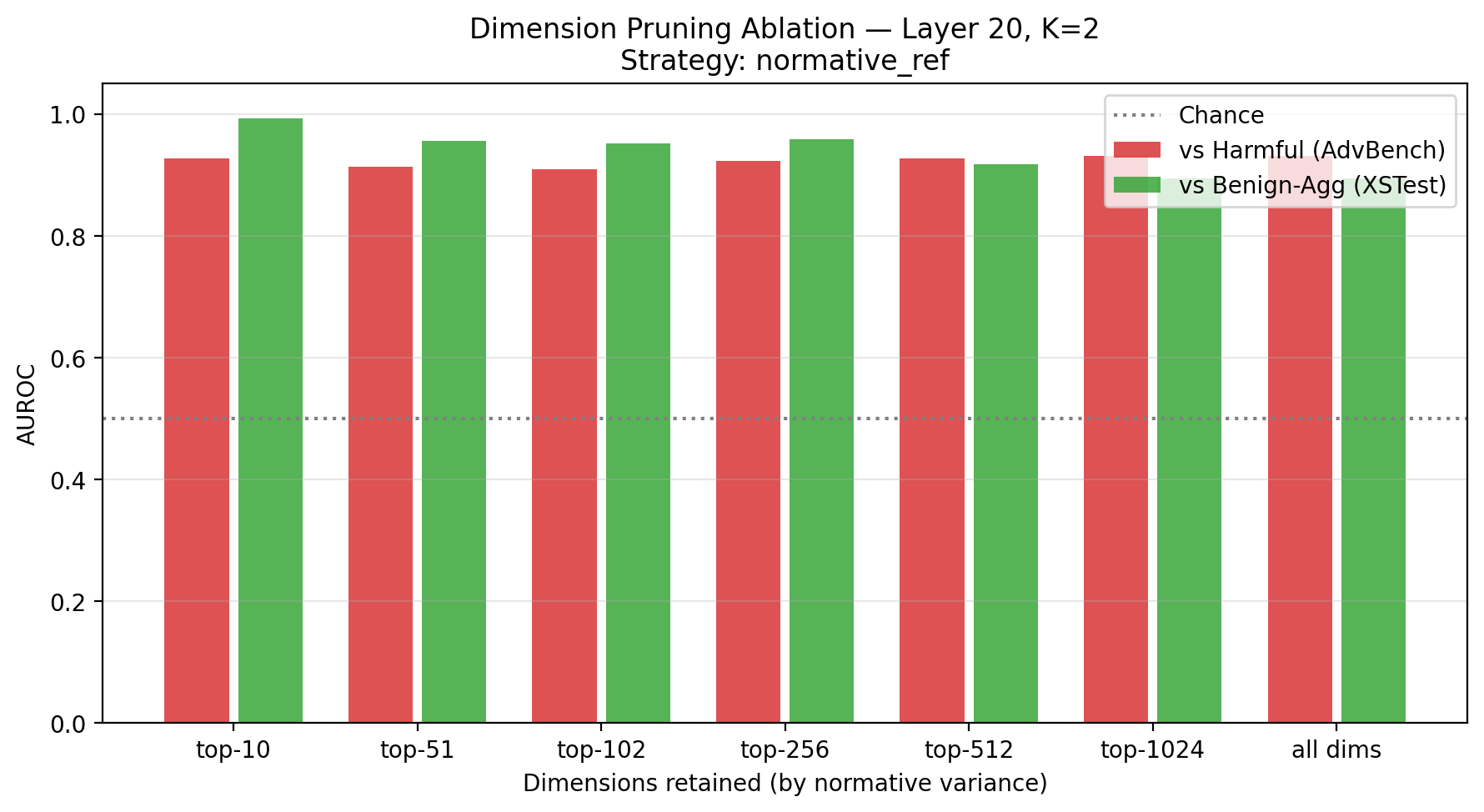}
    \caption{\QABig{}-Base}
  \end{subfigure}\hfill
  \begin{subfigure}[t]{0.48\textwidth}
    \includegraphics[width=\linewidth]{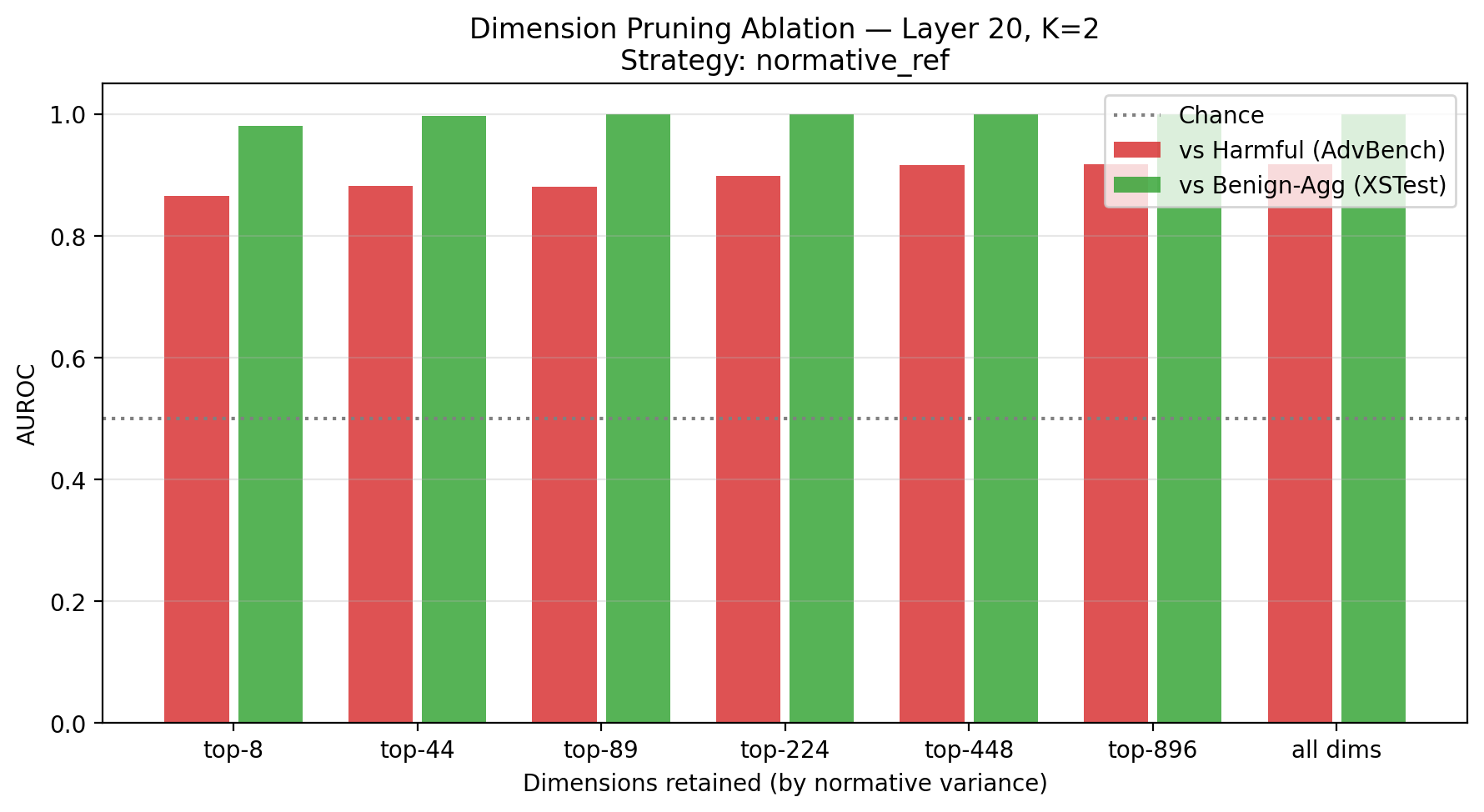}
    \caption{\QASmall{}-Base}
  \end{subfigure}
  \caption{%
    \textbf{Dimension-pruning ablation at the operating layer ($K{=}2$)} for
    the two base variants; all six models are shown in \cref{fig:app_dim_all}.
    AUROC h/n (red) and AUROC h/b (green) vs.\ number of principal dimensions
    retained by descending normative variance.
    Both tasks are maximised at the top-10 dimensions (${\approx}1\%$ of $D$)
    and degrade monotonically thereafter. This pattern indicates that retaining additional
components does not improve performance under this setup.
  }
  \label{fig:ablation_dim}
\end{figure}

\Cref{fig:ablation_dim} shows AUROC as a function of the number of principal
dimensions retained (by descending normative variance), for $K{=}2$ at the
operating layer, for the two base variants.
The full six-model ablation is in \cref{fig:app_dim_all}.
In every model, retaining just the top-10 dimensions (${\approx}1\%$ of
$D{=}1024$ or $1.1\%$ of $D{=}896$) simultaneously maximises both h/n and h/b
AUROC; additional dimensions monotonically dilute performance.
The safety signal is thus concentrated in a compact subspace. This result
is consistent with the near-one-dimensional geometry reported by
\citet{arditi2024refusal} and reinforced here across six model variants.

\subsection{Normative set stability}
\label{sec:stability}

\begin{figure}[htbp]
  \centering
  \begin{subfigure}[t]{1.0\textwidth}
    \includegraphics[width=\linewidth]{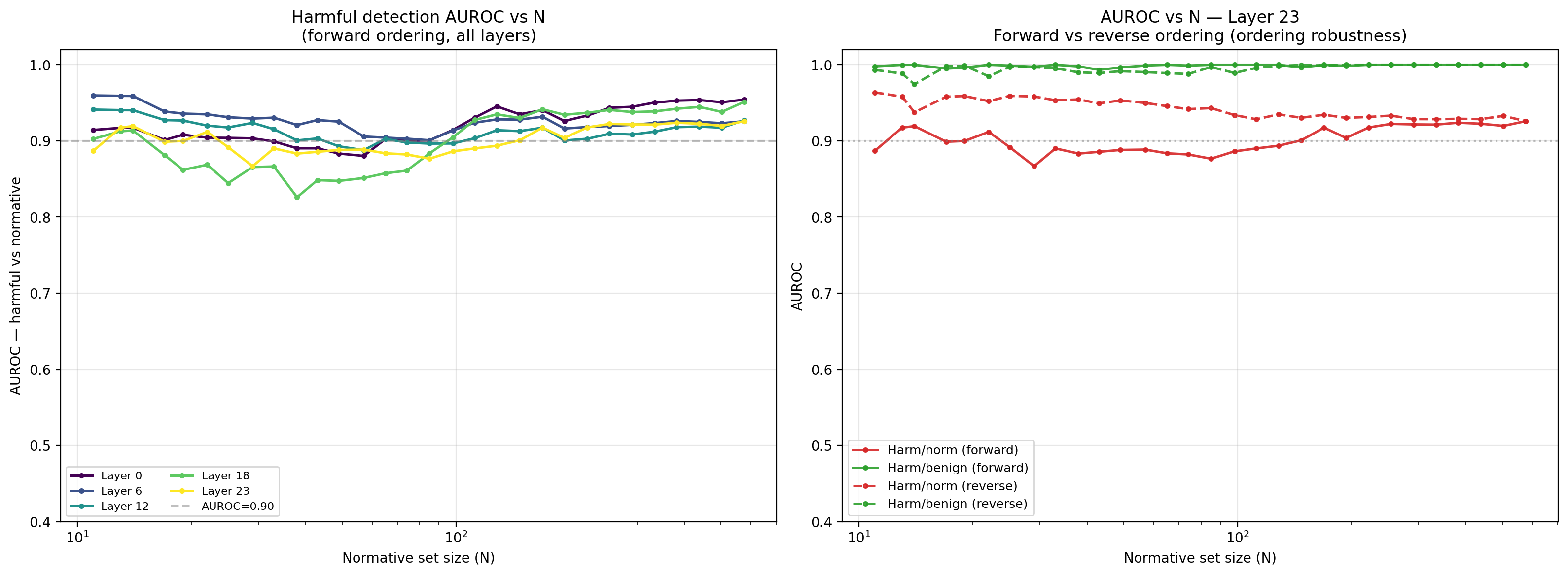}
    \caption{\QABig{}-Base}
  \end{subfigure}
  \\[6pt]
  \begin{subfigure}[t]{1.0\textwidth}
    \includegraphics[width=\linewidth]{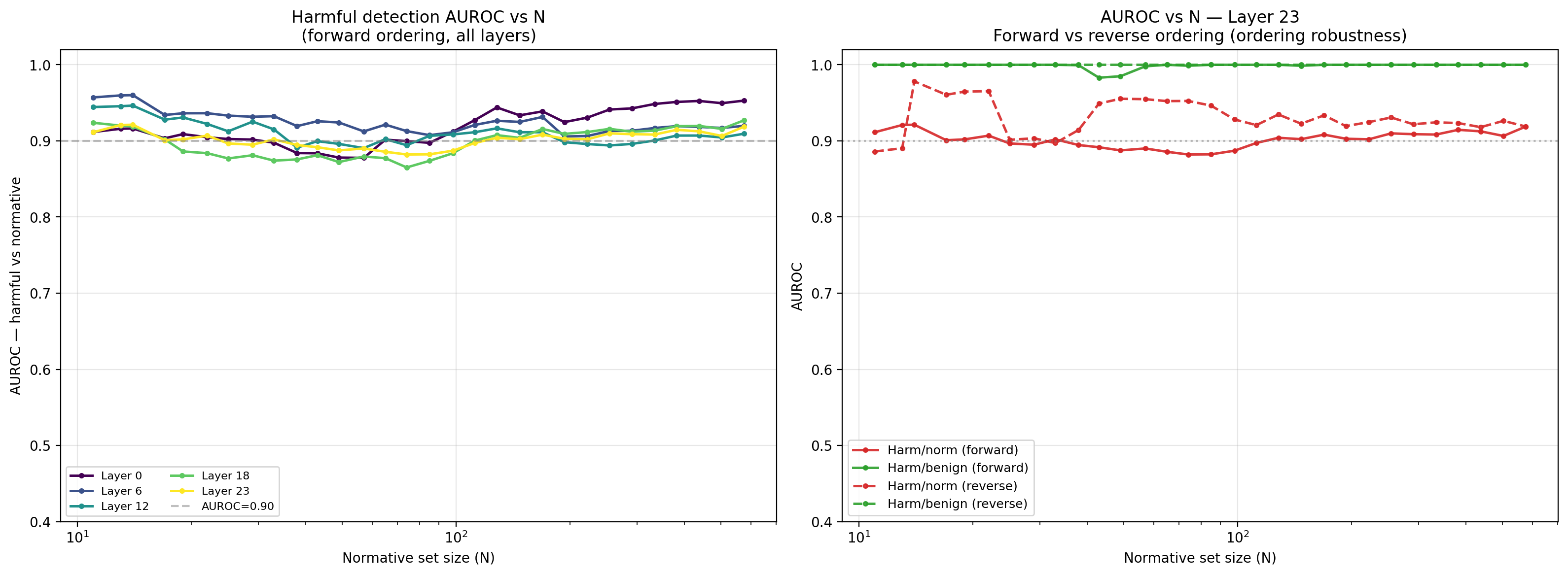}
    \caption{\QABig{}-Chat}
  \end{subfigure}
  \\[6pt]
  \begin{subfigure}[t]{1.0\textwidth}
    \includegraphics[width=\linewidth]{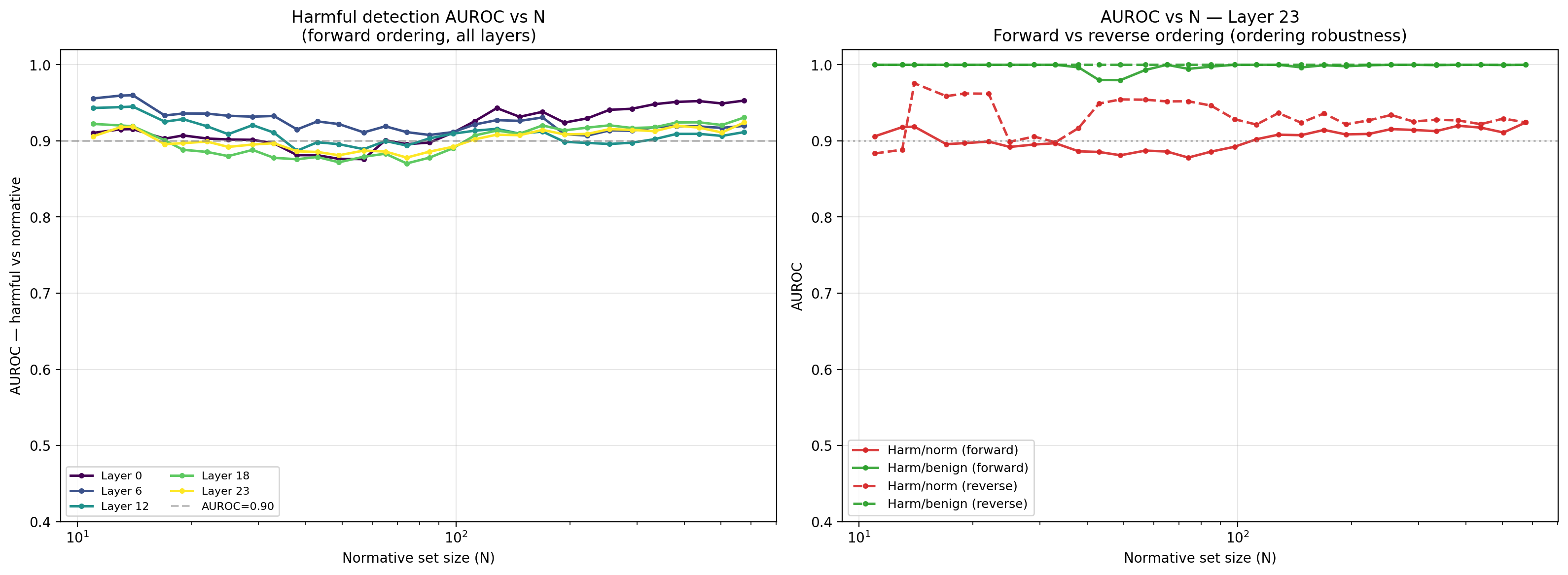}
    \caption{\QABig{}-Abliterated}
  \end{subfigure}
  \caption{%
    \textbf{AUROC vs.\ normative fit-set size $N$ for the \QABig{} family.}
    \textbf{Left sub-panels}: AUROC h/n vs.\ $N$ for representative layers
    (forward ordering), showing stabilisation well before $N{=}200$ at every layer.
    \textbf{Right sub-panels}: AUROC vs.\ $N$ at a fixed late layer, comparing
    forward (solid) and reverse (dashed) ordering for AUROC h/n (red) and
    AUROC h/b (green). Green curves are flat at 1.000 throughout; red curves are 
    stable above 0.90 at all $N{\geq}30$ and invariant to ordering, ruling out 
    sample-ordering artefacts.
  }
  \label{fig:stability_qwen35}
\end{figure}

\begin{figure}[htbp]
  \centering
  \begin{subfigure}[t]{1.0\textwidth}
    \includegraphics[width=\linewidth]{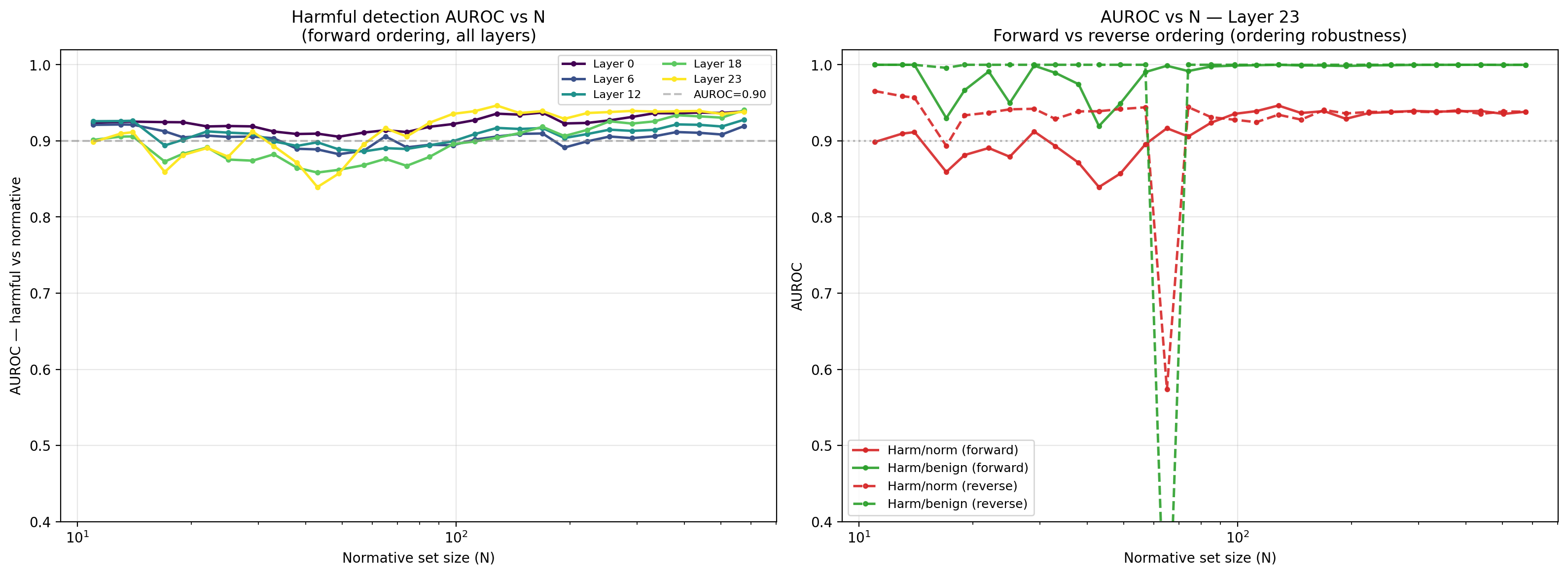}
    \caption{\QASmall{}-Base}
  \end{subfigure}
  \\[6pt]
  \begin{subfigure}[t]{1.0\textwidth}
    \includegraphics[width=\linewidth]{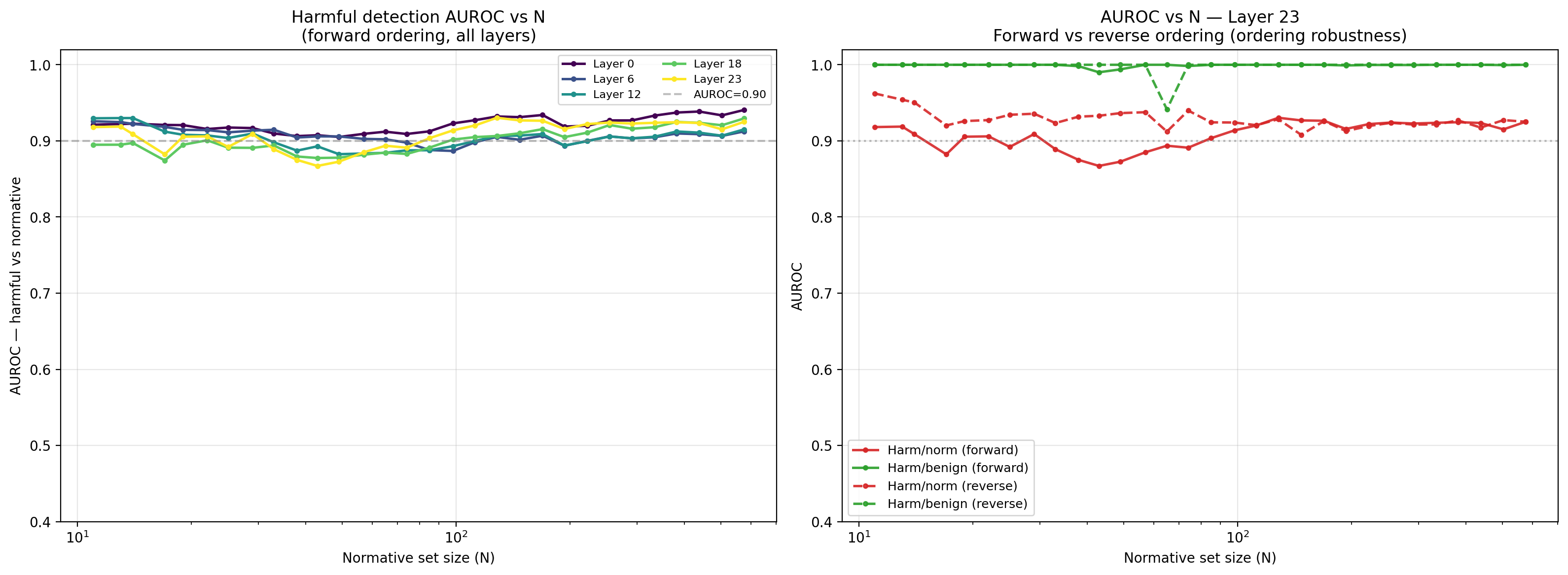}
    \caption{\QASmall{}-Instruct}
  \end{subfigure}
  \\[6pt]
  \begin{subfigure}[t]{1.0\textwidth}
    \includegraphics[width=\linewidth]{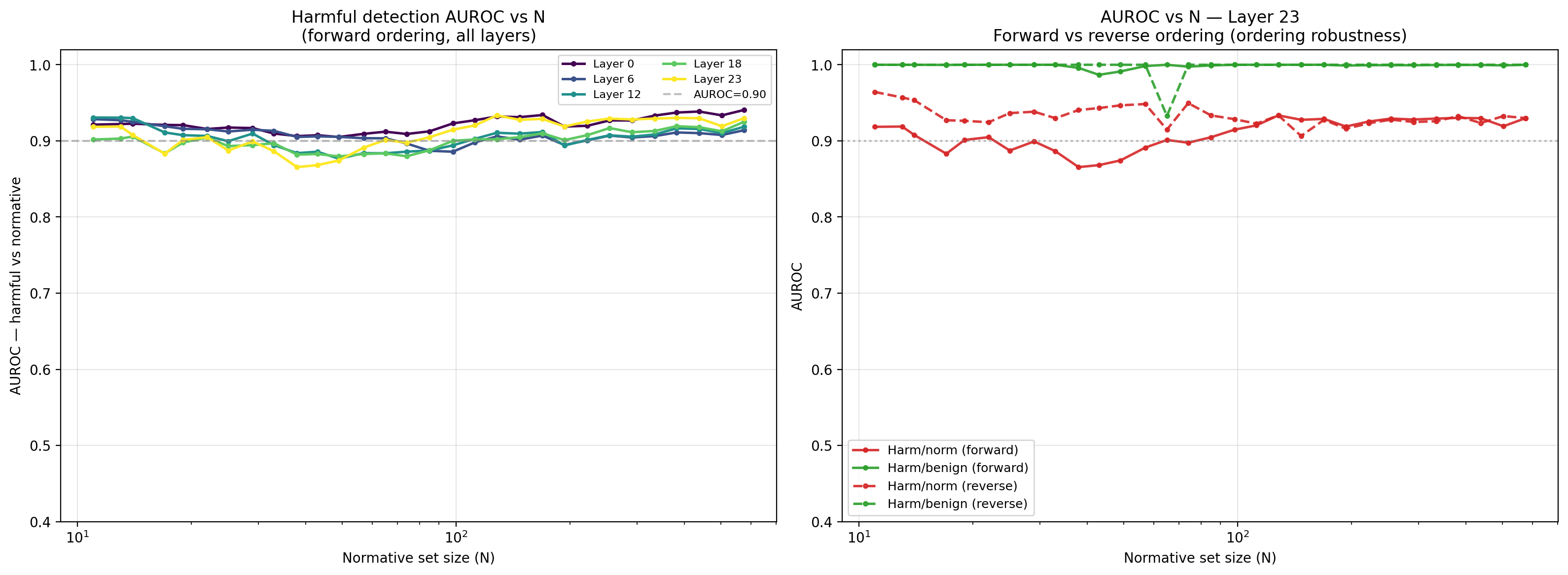}
    \caption{\QASmall{}-Abliterated}
  \end{subfigure}
  \caption{%
    \textbf{AUROC vs.\ normative fit-set size $N$ for the \QASmall{} family.}
    Similar to the 0.8B architecture, performance stabilises well before $N{=}200$ 
    (left panels) and exhibits strict ordering invariance (right panels). 
    The abliterated variant (panel c) shows the same robust stability pattern at 
    its operating layer (layer 10) as at other layers, confirming data-size 
    requirements are unaffected by the ablation of refusal directions.
  }
  \label{fig:stability_qwen25}
\end{figure}

\Cref{fig:stability_qwen35} and \cref{fig:stability_qwen25} plot AUROC as a function of normative fit-set size $N$ for all variants of the \QABig{} and \QASmall{} families, respectively.

Performance saturates remarkably early. Across all models and nearly all layers, AUROC$_{\text{h/n}}$ already exceeds 0.90 once $N\gtrsim 100$, with many layers reaching this threshold throughout most layers. Even with extremely small normative sets ($N=10$--20), late-layer performance remains strong (typically $>0.85$). The harmful-versus-benign-aggressive separation reaches and remains perfectly flat at AUROC$_{\text{h/b}}=1.000$ from the smallest tested values of $N$.

The right sub-panels confirm near-perfect invariance to the ordering of the normative set: forward (solid) and reverse (dashed) curves overlap almost completely after $N\approx 30$. This rules out sample-ordering artefacts and shows that the leading principal component rapidly converges to a stable reference direction.

Notably, the stability profiles are qualitatively consistent across Base, Chat, and Abliterated variants within each family. This provides further evidence that the harmful-intent geometry exploited by LatentBiopsy is formed during pretraining and remains largely unaffected by later instruction tuning or refusal ablation. Several adjacent late layers also show nearly identical curves, indicating that the relevant structure isn't limited to a single layer but is distributed across a small range of layers.

The immediate perfect separability of harmful versus benign-aggressive prompts across all $N$ further suggests that these categories occupy well-separated regions of representation space, independent of the normative reference.

Taken together, these results demonstrate that LatentBiopsy is highly data-efficient: a few dozen safe prompts suffice to construct a high-quality reference direction, and $N=200$ lies comfortably in the saturated regime.

\section{Discussion}
\label{sec:discussion}

\paragraph{The abliteration result and its safety implications.}
The abliterated variants apply targeted removal of the learned refusal
direction, aiming to eliminate refusal-style behaviour.
Yet \method{} achieves AUROC h/b $= 1.000$ and AUROC h/n within 0.005 of the
corresponding instruction-tuned models in both families.
This establishes a \emph{geometric dissociation}: harmful semantic intent is
represented in the residual stream independently of the downstream mechanism
that acts on it.
Recent safety approaches such as \citet{zheng2024prompt} target the refusal
direction to strengthen model behaviour.
Our findings suggest that such interventions address the generative mechanism
without altering the representational geometry; the latent signal persists
even when the direction has been mathematically erased.
This is consequential for both offensive and defensive AI safety: a model that
cannot refuse retains an intact, exploitable signal for an external detector,
and an adversary who abliterates a model to bypass its safeguards does not
thereby erase the geometric evidence of harmful intent.

\paragraph{Why do the two families have opposite ring orientations?}
At layer 20, \QABig{} harmful prompts are more angular from PC1 than normative
prompts ($\Delta\theta = +0.62$--$+0.65$~rad), while \QASmall{} harmful prompts
are more aligned ($\Delta\theta = -0.45$--$-0.60$~rad).
This family-level difference likely reflects how safety-relevant representations
are encoded relative to the dominant normative variance direction in each
architecture, a quantity that depends on pretraining data mixture, model
width, and architectural details simultaneously.
We do not have a mechanistic account and regard the cause as an open question.
What the result does establish is that no single ring orientation can be assumed
\emph{a priori} across architectures, making direction-agnostic scoring a
structural requirement rather than a design choice.

\paragraph{The near-degenerate harmful compactness.}
Across all six models, $\sigma_\theta^\mathrm{harm}\approx0.03$--$0.05$~rad, 
one order of magnitude smaller than $\sigma_\theta^\mathrm{norm}$.
This compactness survives abliteration, ruling out the refusal direction as its
source.
The most parsimonious explanation is that AdvBench prompts share a narrow
syntactic template producing near-identical last-token activations.
Evaluation on structurally diverse datasets such as
JailbreakBench~\citep{chao2024jailbreakbench} is the key open experiment to
determine whether this compactness is a surface-form artefact or a genuine
semantic regularity.
This question is the primary empirical limitation of the present work.

\paragraph{Does safety geometry precede alignment?}
Base models achieve at least equal harmful-detection AUROC as their
instruction-tuned counterparts in both families, and the two-ring structure is
qualitatively identical across base, instruct, and abliterated variants.
This is consistent with the hypothesis that harmful-intent geometry is
established during pretraining and is not a product of alignment fine-tuning.
However, the observation is currently limited to two model families from a
single vendor, and Qwen-specific data mixture effects cannot be ruled out.
Extending to at least one non-Qwen family is
the highest-priority next experiment.

\paragraph{Limitations and open directions.}
\textit{Cross-architecture validation} is the most critical gap: the
geometry-precedes-alignment and geometry-survives-ablation findings must be
replicated in at least one non-Qwen family before they can be treated as
general results.
\textit{Dataset diversity}: evaluation on JailbreakBench will determine whether
harmful compactness generalises beyond the lexically homogeneous AdvBench.
\textit{Adversarial robustness}: prompts crafted to minimise $|s(x)|$ while
preserving harmful intent are the natural white-box attack on \method{}; their
effect is entirely untested and constitutes the most important open safety
question.
\textit{Layer selection}: a dedicated layer-selection split would eliminate
the one level of selection bias and yield strictly unbiased AUROC estimates.
\textit{Calibrated thresholding}: deployment requires a principled approach to
threshold selection given the variability of benign-aggressive placement across
families.

\section{Conclusion}
\label{sec:conclusion}

\method{} demonstrates that a direction-agnostic angular anomaly detector,
built exclusively from 200 safe activations, robustly identifies harmful prompts
across base, instruction-tuned, and abliterated variants of two model families.
The method achieves harmful versus normative AUROC $\geq$0.937 and harmful versus benign-aggressive AUROC $= 1.000$ across all
six tested variants, with sub-millisecond per-query overhead and no harmful
training data.

The central finding is a \emph{geometric dissociation}: removing the refusal
mechanism from a model in both a 0.8B and a 0.5B architecture leaves
harmful-intent geometry intact.
A model that cannot refuse retains the latent signal exploitable by an external
detector, and an adversary who abliterates refusal does not thereby erase the
evidence.
Combined with the observation that base models match instruction-tuned ones
on all detection metrics, the evidence is consistent with harmful-intent
geometry being established during pretraining, independently of the alignment
process in both its presence and its absence.

An unexpected empirical pattern (though not extensively tested) is the opposite ring orientation across
families: harmful prompts are the most angular group in \QABig{} and the most
aligned group in \QASmall{}.
The anomaly score handles both without modification.
Explaining this architectural difference mechanistically, and confirming whether
the pattern generalises beyond the Qwen family, is the natural next step of this
research programme.


\section{Ethical Considerations and Broader Impact}
\label{sec:ethics}

\method{} provides a diagnostic capability to identify harmful instructions at inference time. While the primary objective is to enhance safety, we acknowledge the inherent dual-use potential of interpretability tools. 

\paragraph{Responsible Research Practice.}
We have utilized only publicly available datasets (AdvBench, XSTest, Alpaca) that are standard within the AI safety literature. We do not provide, encourage, or facilitate the generation of new harmful content. We strictly adhere to the AI safety community's norms of responsible disclosure and advocate for the use of latent analysis solely to improve model robustness, interpretability, and safety alignment. We believe that democratising the ability to "read" harmful intent is a net positive for safety, as it reduces the reliance on "black-box" proprietary safety filters and enables transparent, model-agnostic verification of safety alignment.

\paragraph{Code and Data Availability.}

All code is available at \url{https://github.com/isaac-6/geometric-latent-biopsy}.
A Zenodo archive is at \url{https://doi.org/10.5281/zenodo.19294977}.
Datasets: Alpaca-Cleaned~\citep{taori2023alpaca},
AdvBench~\citep{zou2023universal},
XSTest~\citep{rottger2023xstest}.

Models used in this work:
Qwen/Qwen3.5-0.8B-Base,
Qwen/Qwen3.5-0.8B,
prithivMLmods/Gliese-Qwen3.5-0.8B-Abliterated-Caption,
Qwen/Qwen2.5-0.5B,
Qwen/Qwen2.5-0.5B-Instruct,
huihui-ai/Qwen2.5-0.5B-Instruct-abliterated.

\bibliography{references}

@misc{zou2023universal,
  title        = {Universal and Transferable Adversarial Attacks on Aligned Language Models},
  author       = {Zou, Andy and Wang, Zifan and Kolter, J. Zico and Fredrikson, Matt},
  howpublished = {arXiv preprint arXiv:2307.15043},
  year         = {2023}
}

@misc{zou2023representation,
  title        = {Representation Engineering: A Top-Down Approach to {AI} Transparency},
  author       = {Zou, Andy and Phan, Long and Chen, Sarah and Campbell, James and
                  Guo, Phillip and Ren, Richard and Pan, Alexander and Yin, Xuwang and
                  Mazeika, Mantas and Dombrowski, Ann-Kathrin and others},
  howpublished = {arXiv preprint arXiv:2310.01405},
  year         = {2023}
}

@misc{arditi2024refusal,
  title        = {Refusal in Language Models Is Mediated by a Single Direction},
  author       = {Arditi, Andy and Obeso, Oscar and Syed, Aaquib and Cunningham, Hoagy and
                  Filan, Daniel and Colognese, Fabien and Wattenberg, Martin and
                  Vi{\'e}gas, Fernanda},
  howpublished = {arXiv preprint arXiv:2406.11717},
  year         = {2024}
}

@misc{inan2023llama,
  title        = {Llama Guard: {LLM}-based Input-Output Safeguard for Human-{AI} Conversations},
  author       = {Inan, Hakan and Upasani, Kartikeya and Chi, Jianfeng and Rungta, Rashi and
                  Iyer, Krithika and Mao, Yuning and Tontchev, Michael and Hu, Qing and
                  Fuller, Brian and Testuggine, Davide and Khabsa, Madian},
  howpublished = {arXiv preprint arXiv:2312.06674},
  year         = {2023}
}

@misc{rottger2023xstest,
  title        = {{XSTest}: A Test Suite for Identifying Exaggerated Safety Behaviours
                  in Large Language Models},
  author       = {R{\"o}ttger, Paul and Kirk, Hannah Rose and Vidgen, Bertie and
                  Attanasio, Giuseppe and Bianchi, Federico and Hovy, Dirk},
  howpublished = {arXiv preprint arXiv:2308.01263},
  year         = {2023}
}

@misc{jain2023baseline,
  title        = {Baseline Defenses for Adversarial Attacks Against Aligned Language Models},
  author       = {Jain, Neel and Schwarzschild, Avi and Wen, Yuxin and Somepalli, Gowthami and
                  Kirchenbauer, John and Chiang, Ping-yeh and Goldblum, Micah and
                  Goldstein, Tom and others},
  howpublished = {arXiv preprint arXiv:2309.00614},
  year         = {2023}
}

@misc{alon2023detecting,
  title        = {Detecting Language Model Attacks with Perplexity},
  author       = {Alon, Gabriel and Kamfonas, Michael},
  howpublished = {arXiv preprint arXiv:2308.14132},
  year         = {2023}
}

@inproceedings{zheng2024prompt,
  title     = {On Prompt-Driven Safeguarding for Large Language Models},
  author    = {Zheng, Chujie and Fan, Lifeng and Chen, Hang and Liu, Yue and
               Huang, Minlie},
  booktitle = {Proceedings of the 41st International Conference on Machine Learning},
  year      = {2024},
  note      = {arXiv preprint arXiv:2401.18018}
}

@article{llorente2025theta,
  title   = {Theta, a multidimensional ratio biomarker applied to five amyloid
             beta peptides for investigations in familial {Alzheimer's} disease},
  author  = {Llorente-Saguer, Isaac and Arber, Charles and Oxtoby, Neil P.},
  journal = {medRxiv},
  year    = {2025},
  publisher={Cold Spring Harbor Laboratory Press},
  note    = {Preprint. \textbf{https://doi.org/10.1101/2025.08.06.25333131}}
}

@inproceedings{xiong2020layer,
  title={On layer normalization in the transformer architecture},
  author={Xiong, Ruibin and Yang, Yunchang and He, Di and Zheng, Kai and Zheng, Shuxin and Xing, Chen and Zhang, Huishuai and Lan, Yanyan and Wang, Liwei and Liu, Tieyan},
  booktitle={International conference on machine learning},
  pages={10524--10533},
  year={2020},
  organization={PMLR}
}

@article{park2023linear,
  title={The linear representation hypothesis and the geometry of large language models},
  author={Park, Kiho and Choe, Yo Joong and Veitch, Victor},
  journal={arXiv preprint arXiv:2311.03658},
  year={2023}
}

@misc{taori2023alpaca,
  title={Stanford Alpaca: An Instruction-Following {LLaMA} Model},
  author={Taori, Rohan and Gulrajani, Ishaan and Zhang, Tianyi and Dubois, Yann and Li, Xuechen and Guestrin, Carlos and Liang, Percy and Hashimoto, Tatsunori B.},
  howpublished={\url{https://github.com/tatsu-lab/stanford_alpaca}},
  year={2023}
}

@inproceedings{li2023inference,
  title={Inference-Time Intervention: Eliciting Truthful Answers from a Language Model},
  author={Li, Kenneth and Patel, Oam and Vi{\'e}gas, Fernanda and Pfister, Hanspeter and Wattenberg, Martin},
  booktitle={Advances in Neural Information Processing Systems},
  volume={36},
  year={2023}
}

@article{chao2024jailbreakbench,
  title={{JailbreakBench}: An Open Robustness Benchmark for Jailbreaking Large Language Models},
  author={Chao, Patrick and Debenedetti, Edoardo and Robey, Alexander and Andriushchenko, Maksym and Croce, Francesco and Sehwag, Vikash and Dobriban, Edgar and Flammarion, Nicolas and Pappas, George J and Tramer, Florian and others},
  journal={arXiv preprint arXiv:2404.01318},
  year={2024}
}
\bibliographystyle{plainnat}

\appendix

\section{Harmful-Reference Strategy}
\label{app:harmful_ref}

A supervised variant (\emph{harmful-ref}) fits the biomarker on harmful examples
instead of normative ones, achieving comparable detection performance (AUROC
$0.945$--$0.953$ on harmful-vs-normative) but requiring harmful data at fit
time.
All harmful-ref scores require auto-orientation (sign flip) at every layer and
model, because the harmful distribution is so diffuse that held-out harmful
prompts score as more anomalous under their own Gaussian fit than normative
prompts do.
This confirms the compact/diffuse asymmetry at the distributional level: the
harmful manifold has no compact geometric centre that generalises across samples.
Full harmful-ref figures and statistics are available in the project repository.

\newpage
\section{Precision-Recall Curves: All Models}
\label{app:pr_all}

\begin{figure}[htbp]
  \centering
  
  
  \begin{subfigure}[t]{0.48\textwidth}
    \includegraphics[width=\linewidth]{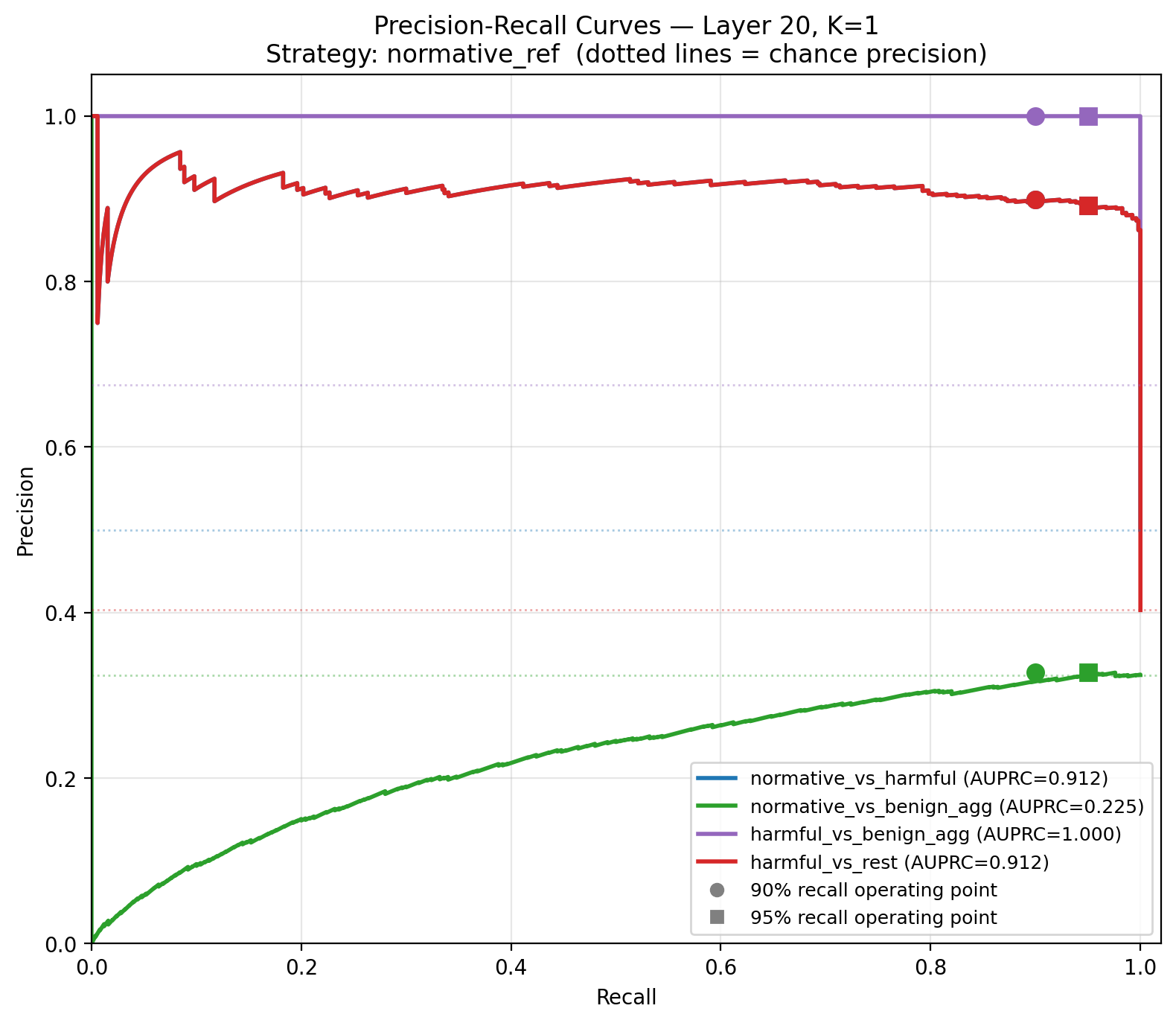}
    \caption{\QABig{}-Chat}
  \end{subfigure}\hfill
  \begin{subfigure}[t]{0.48\textwidth}
    \includegraphics[width=\linewidth]{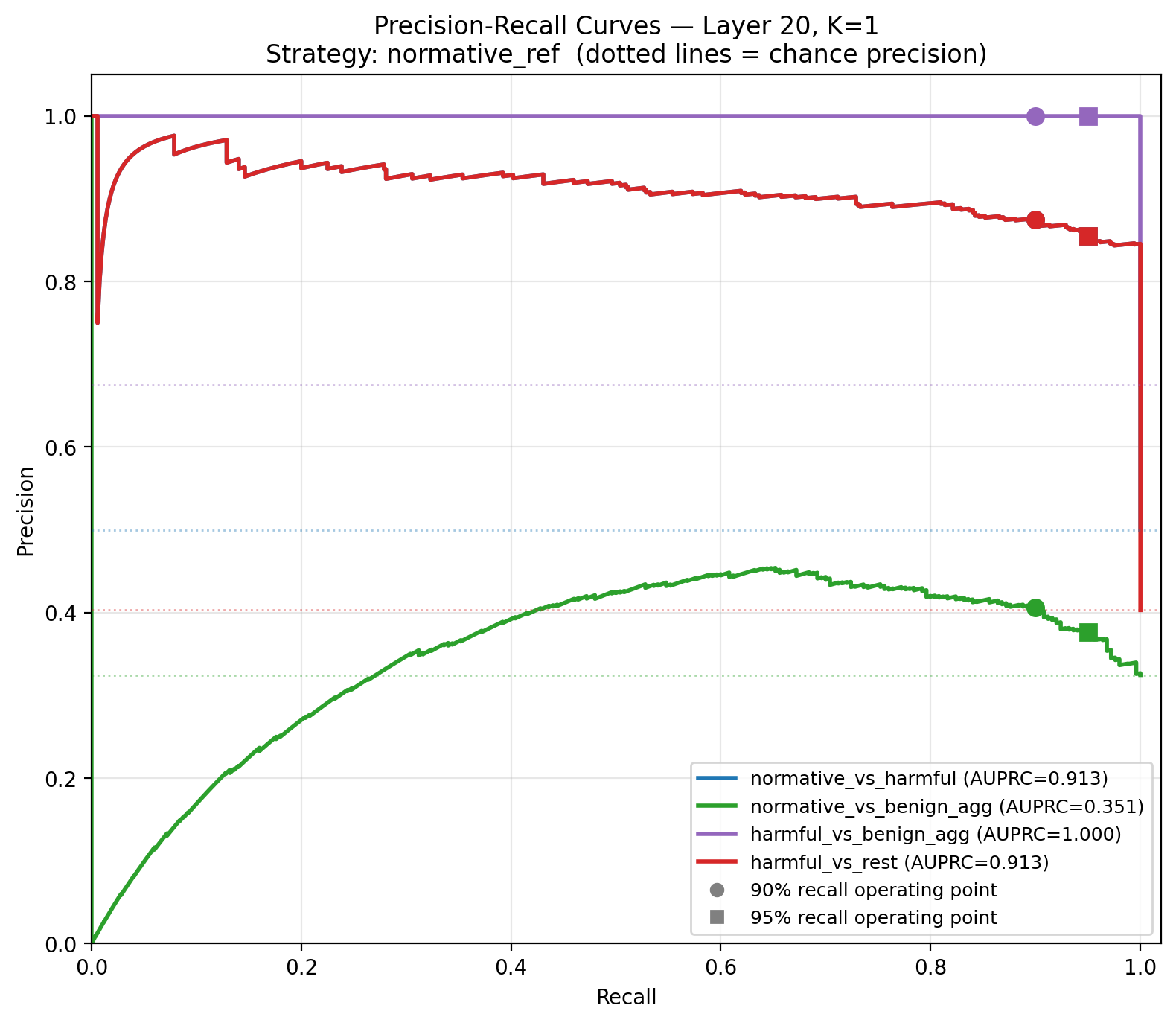}
    \caption{\QASmall{}-Instruct}
  \end{subfigure}
  \\[8pt]
  
  \begin{subfigure}[t]{0.48\textwidth}
    \includegraphics[width=\linewidth]{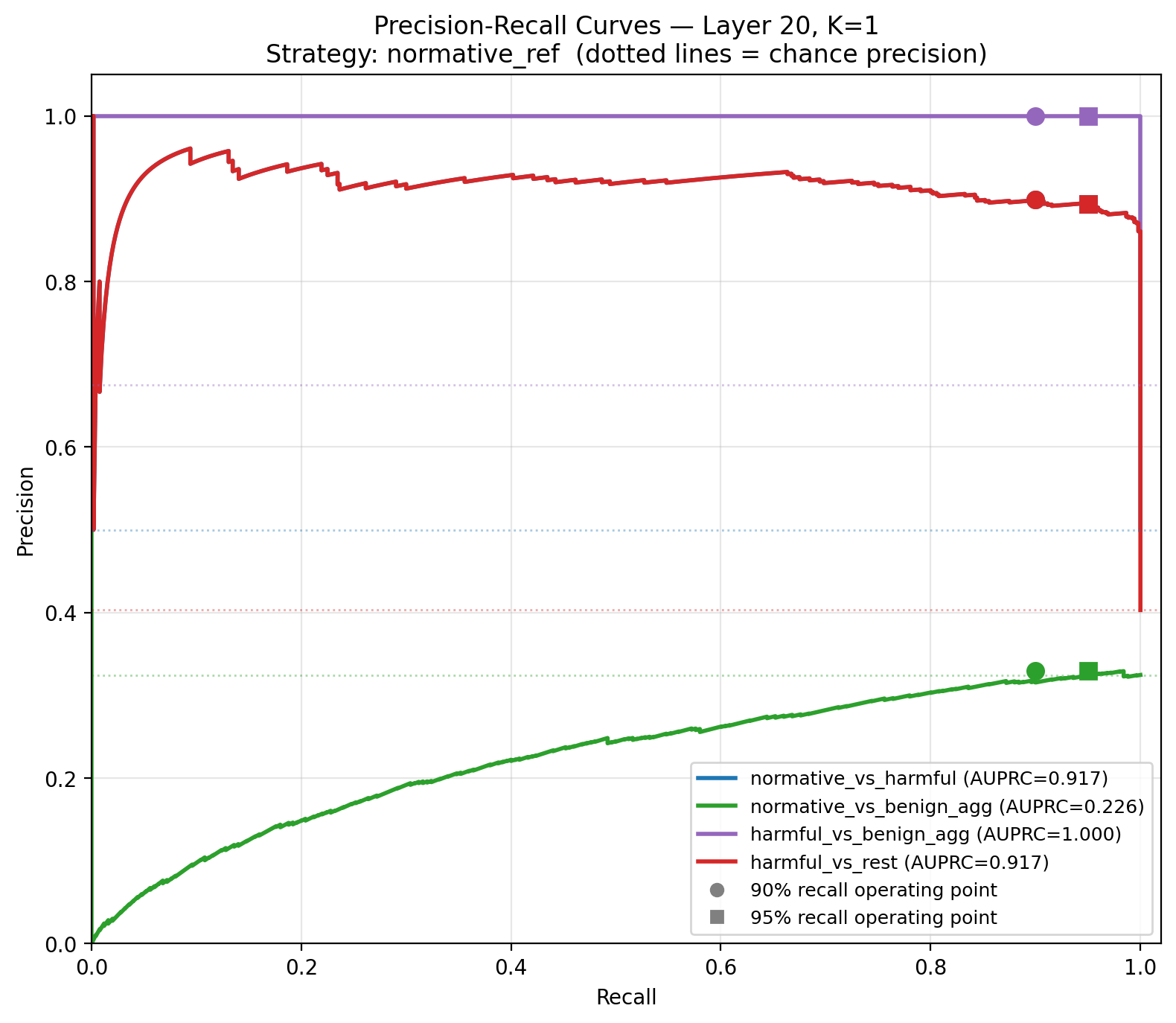}
    \caption{\QABig{}-Abliterated}
  \end{subfigure}\hfill
  \begin{subfigure}[t]{0.48\textwidth}
    \includegraphics[width=\linewidth]{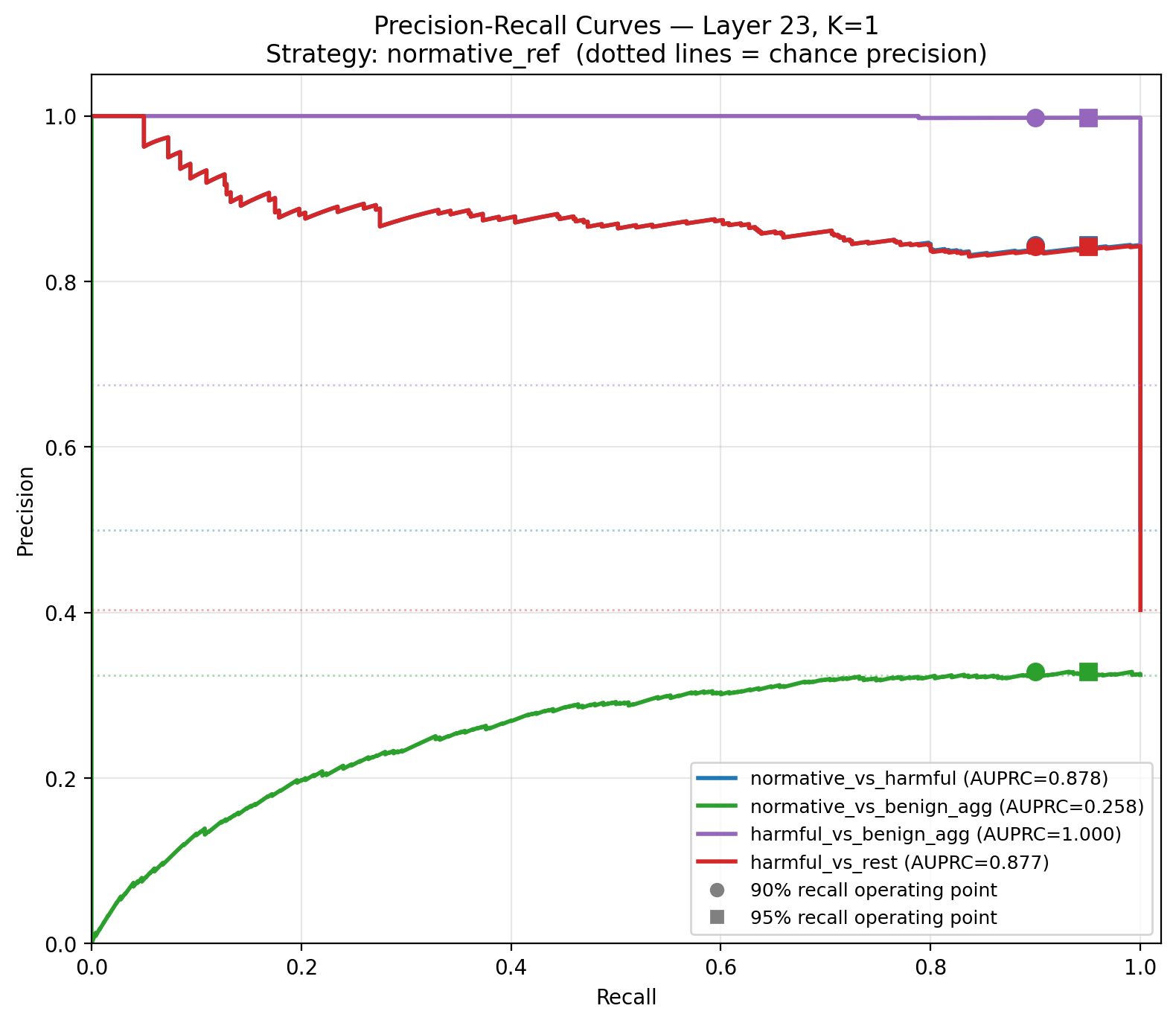}
    \caption{\QASmall{}-Abliterated}
  \end{subfigure}
  
  \caption{Precision-recall curves for all six models; see \cref{fig:pr_curve}
  for a detailed description, and the base models' curves.
  In the \QABig{} family (left column), the normative-vs-benign-agg curve (green) lies below
  chance, confirming that benign-agg prompts are scored as \emph{less} anomalous.
  In the \QASmall{} family (right column), the same curve sits near or slightly above chance,
  reflecting $r_\mathrm{b/n} \approx +0.15$ to $+0.22$.
  In all panels, the harmful-vs-benign-agg curve is flat at precision $= 1.000$.}
  \label{fig:app_pr_all}
\end{figure}

\newpage
\section{Dimension Ablation: All Models}
\label{app:dim_all}

\begin{figure}[htbp]
  \centering
  
  \begin{subfigure}[t]{0.48\textwidth}
    \includegraphics[width=\linewidth]{figures/Qwen3.5-0.8B-Base/auroc_ablation_dim.png}
    \caption{\QABig{}-Base}
  \end{subfigure}\hfill
  \begin{subfigure}[t]{0.48\textwidth}
    \includegraphics[width=\linewidth]{figures/Qwen2.5-0.5B/auroc_ablation_dim.png}
    \caption{\QASmall{}-Base}
  \end{subfigure}
  \\[8pt]
  
  \begin{subfigure}[t]{0.48\textwidth}
    \includegraphics[width=\linewidth]{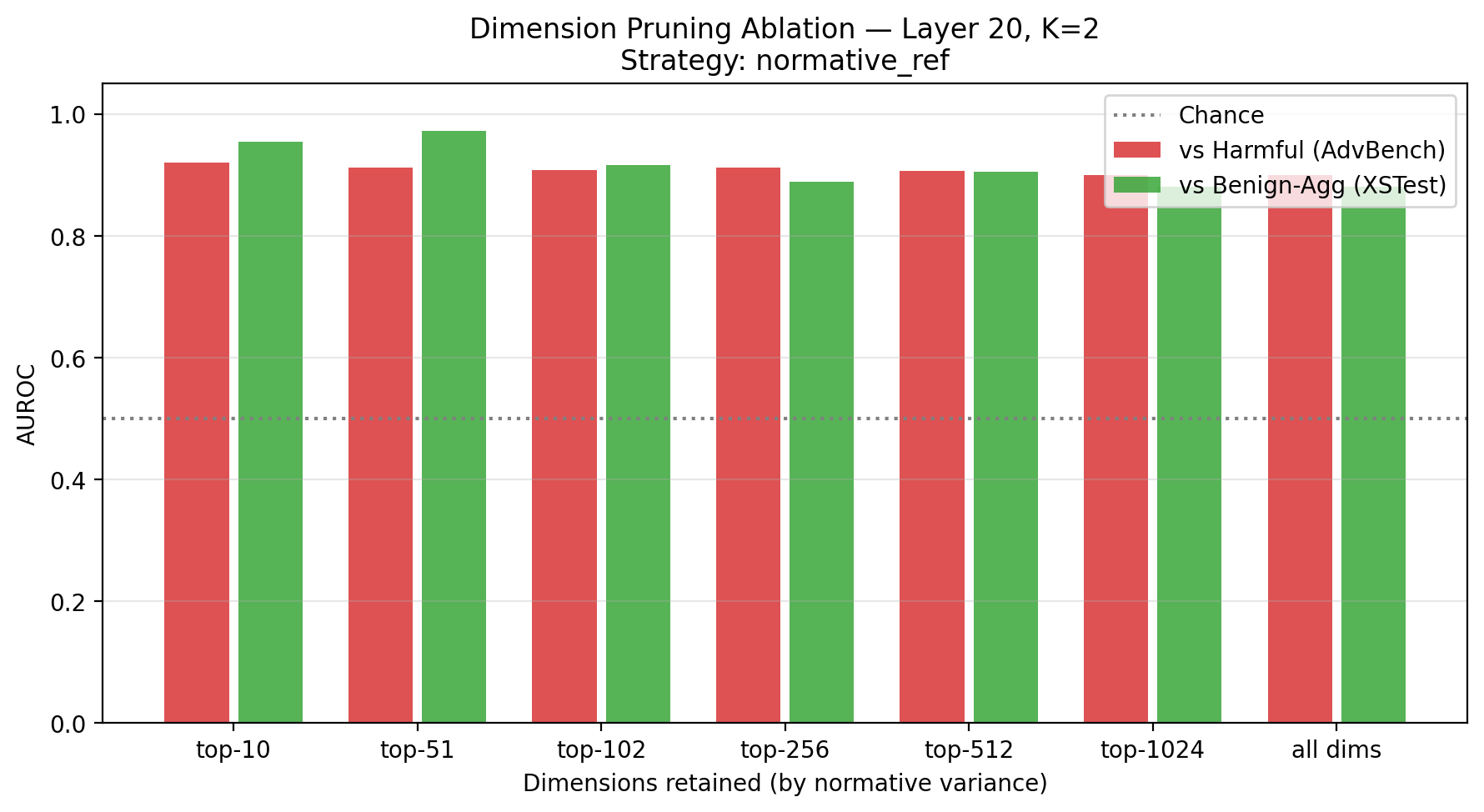}
    \caption{\QABig{}-Chat}
  \end{subfigure}\hfill
  \begin{subfigure}[t]{0.48\textwidth}
    \includegraphics[width=\linewidth]{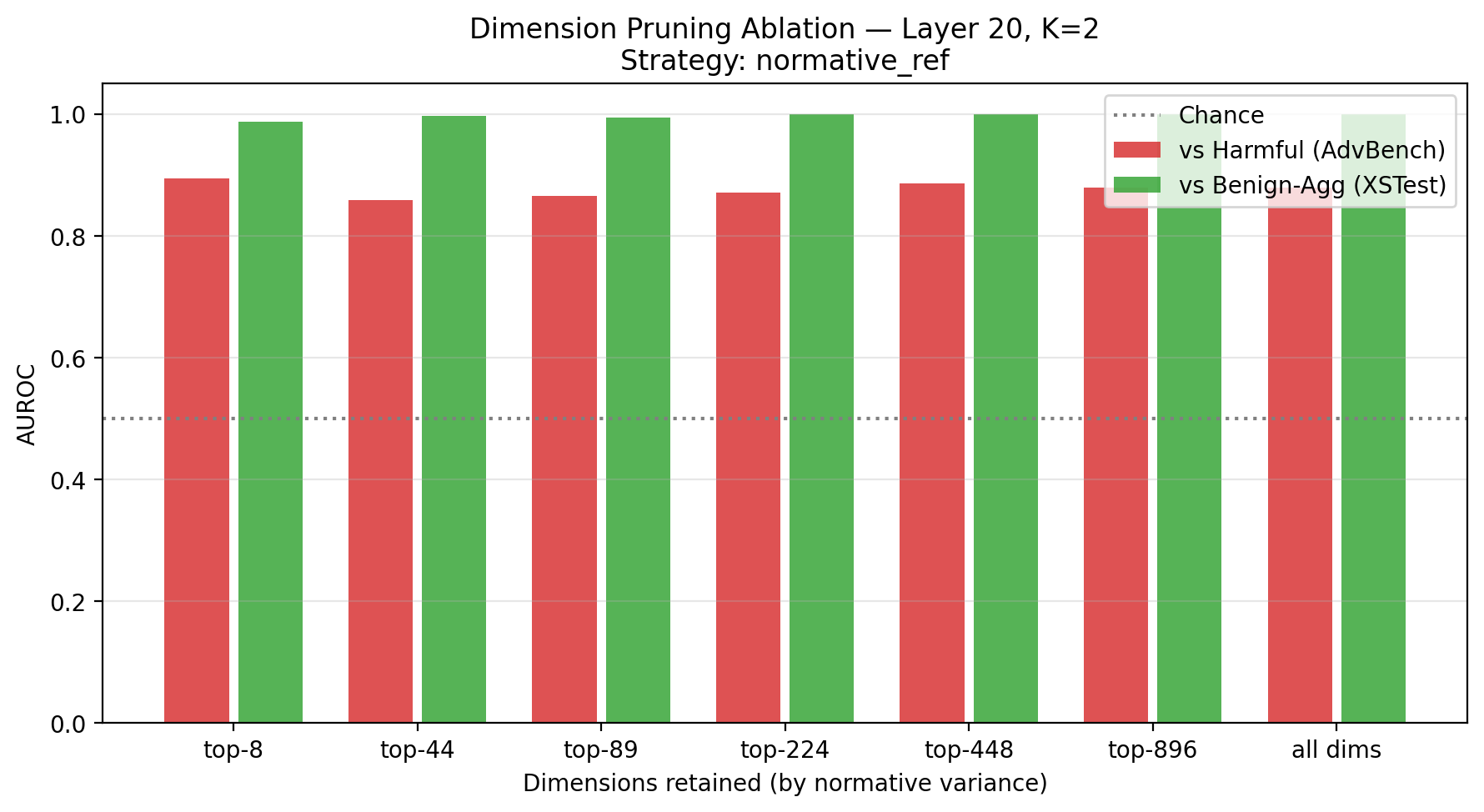}
    \caption{\QASmall{}-Instruct}
  \end{subfigure}
  \\[8pt]
  
  \begin{subfigure}[t]{0.48\textwidth}
    \includegraphics[width=\linewidth]{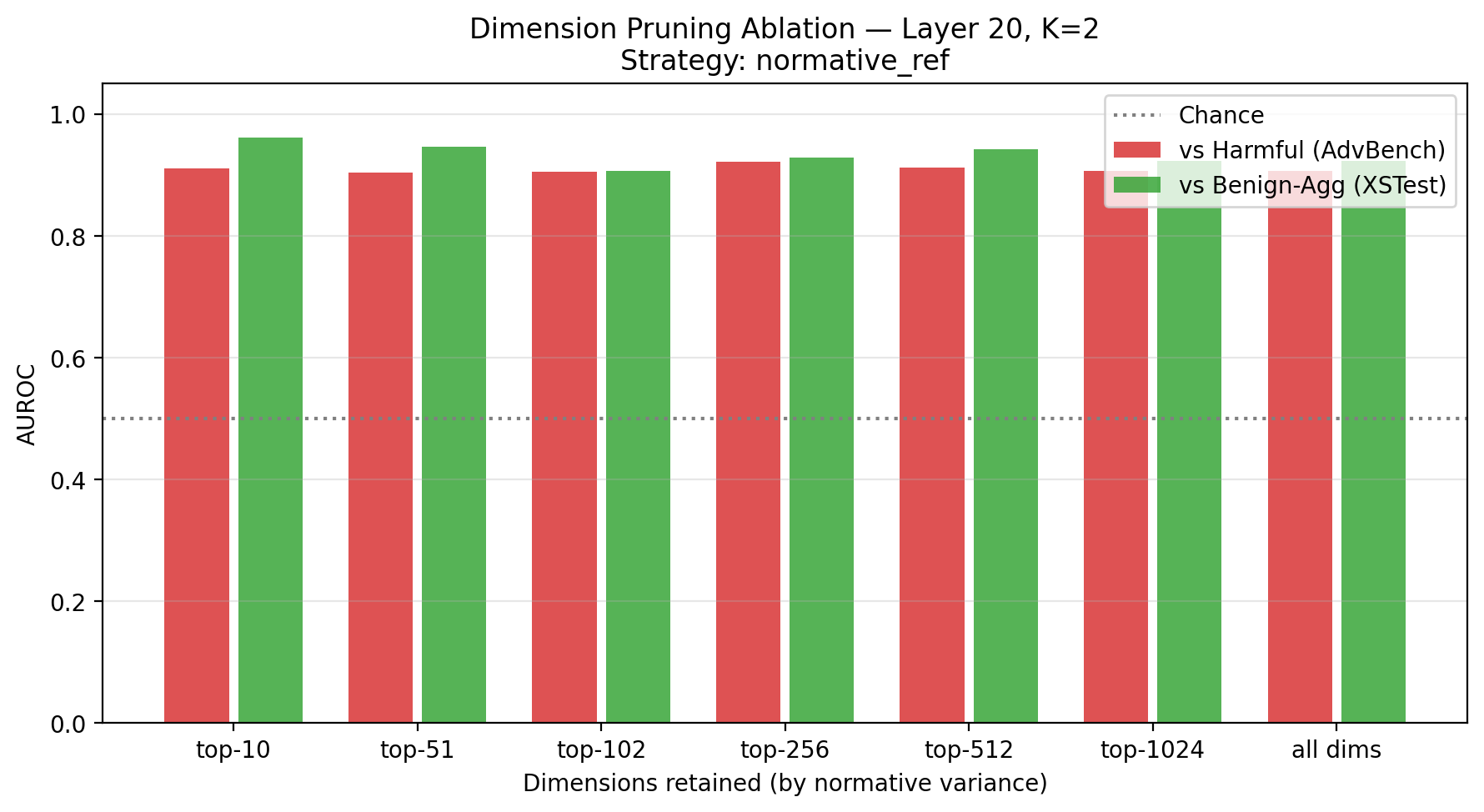}
    \caption{\QABig{}-Abliterated}
  \end{subfigure}\hfill
  \begin{subfigure}[t]{0.48\textwidth}
    \includegraphics[width=\linewidth]{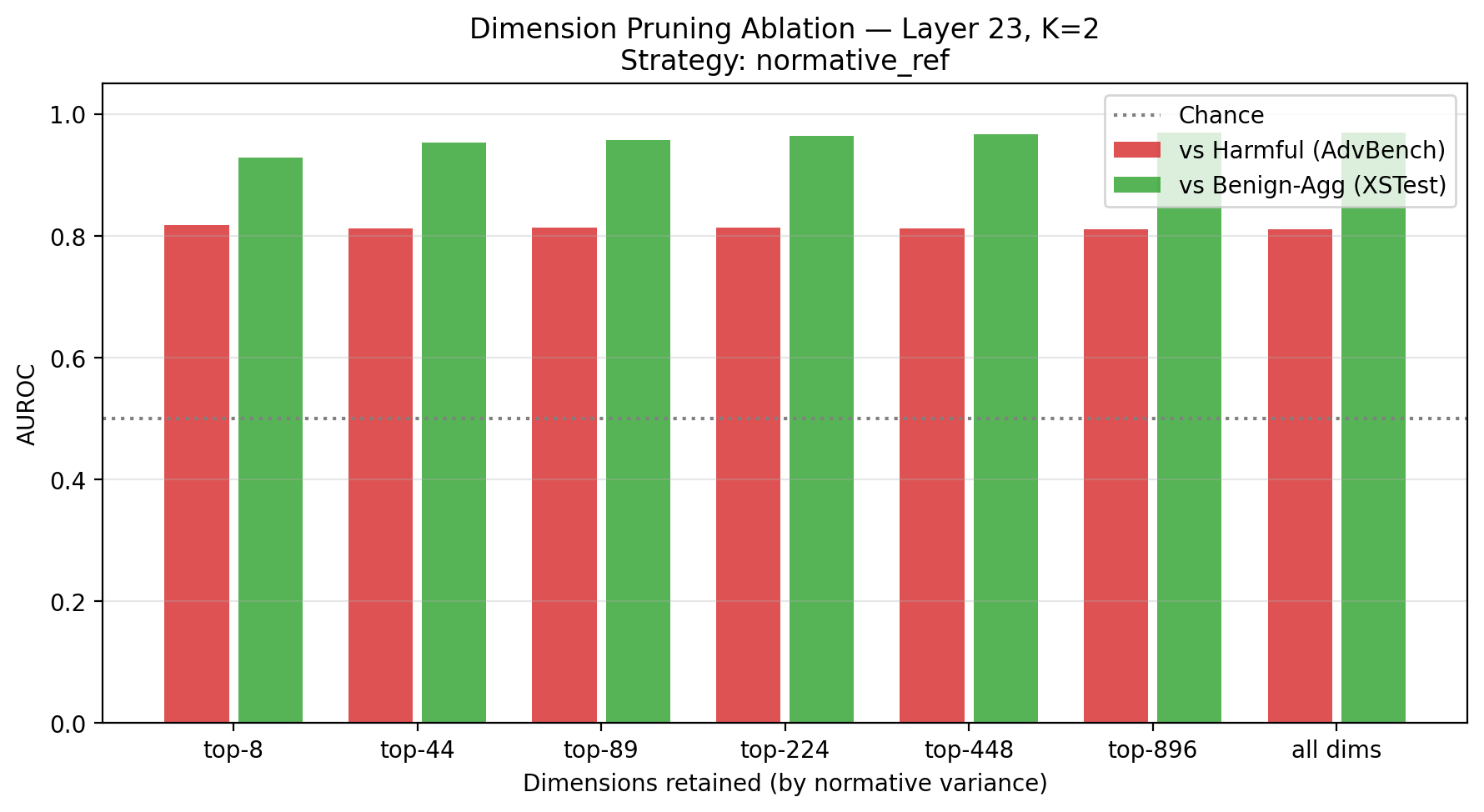}
    \caption{\QASmall{}-Abliterated}
  \end{subfigure}
  
  \caption{Dimension-pruning ablation ($K{=}2$) for all six models;
  see \cref{fig:ablation_dim} for a detailed description.
  In every panel, performance is maximised at top-10 dimensions and
  monotonically decreases thereafter, confirming that the safety signal is
  concentrated in $<1.2\%$ of all residual-stream dimensions across both
  model families.}
  \label{fig:app_dim_all}
\end{figure}

\end{document}